\documentclass{article}
\usepackage{iclr2027_conference,times}

\usepackage{tikz,amssymb}
\usetikzlibrary{arrows.meta}

\RequirePackage{xcolor}
\usepackage{tikz}
\usepackage{graphicx}
\usepackage[normalem]{ulem} %
\usepackage{svg}
\newif\ificlrtodos
\iclrtodostrue          %

\RequirePackage{algorithm}
\RequirePackage{algpseudocode}

\usepackage{amsmath,amsfonts,bm}

\def\eqref#1{equation~\ref{#1}}

\def\1{\bm{1}}

\DeclareMathAlphabet{\mathsfit}{\encodingdefault}{\sfdefault}{m}{sl}
\SetMathAlphabet{\mathsfit}{bold}{\encodingdefault}{\sfdefault}{bx}{n}

\usepackage{hyperref}
\hypersetup{hidelinks}
\usepackage{url}
\usepackage{cleveref}
\makeatletter
\providecommand{\theHALG@line}{\thealgorithm.\arabic{ALG@line}}
\makeatother
\newcounter{tablenote}[table]
\renewcommand{\thetablenote}{\alph{tablenote}}
\crefname{definition}{Def.}{Defs.}
\Crefname{definition}{Def.}{Defs.}
\usepackage{array,booktabs,float}
\usepackage{etoc}

\crefname{figure}{Fig.}{Figs.}
\Crefname{figure}{Fig.}{Figs.}
\crefname{table}{Tab.}{Tabs.}
\Crefname{table}{Tab.}{Tabs.}
\crefname{section}{Sec.}{Secs.}
\Crefname{section}{Sec.}{Secs.}
\crefname{equation}{eq.}{eqs.}
\Crefname{equation}{Eq.}{Eqs.}
\crefname{proposition}{Prop.}{Props.}
\Crefname{proposition}{Prop.}{Props.}
\crefname{theorem}{Thm.}{Thms.}
\Crefname{theorem}{Thm.}{Thms.}
\crefname{corollary}{Cor.}{Cors.}
\Crefname{corollary}{Cor.}{Cors.}
\crefname{appendix}{App.}{Apps.}
\Crefname{appendix}{App.}{Apps.}
\Crefname{algorithm}{Alg.}{Algs.}
\AddToHook{cmd/appendix/before}{%
    \crefalias{section}{appendix}%
    \crefalias{subsection}{appendix}%
}

\newcommand{\method}{Wasserstein Sampling via Bridge Interpolants}

\newcommand{\methodshort}
{WASABI}%
\definecolor{wasabigreen}{HTML}{347A46}
\newcommand{\wasabi}{\textcolor{wasabigreen}{\methodshort}}

\usepackage{amsthm}

\theoremstyle{plain}
\newtheorem{theorem}{Theorem}[section]
\newtheorem{lemma}[theorem]{Lemma}
\newtheorem{proposition}[theorem]{Proposition}
\newtheorem{corollary}[theorem]{Corollary}
\theoremstyle{definition}
\newtheorem{definition}[theorem]{Definition}
\theoremstyle{remark}

\title{WASABI: \method}
\author{%
  Sebastian Salazar$^{1,2}$\thanks{Corresponding authors:
    \texttt{ss5971@columbia.edu} and
    \texttt{ lwu86@jh.edu }.}\\
  \And
  Dilum Fernando$^{3}$\\
  \And
  Michal Kucer$^{2}$\\
  \AND
  Emily Casleton$^{2}$\\
  \And
  Luhuan Wu$^{3*}$\thanks{Equal advising.}\\
  \And
  David M. Blei$^{1\dagger}$\\
  \AND
  {\normalfont\small
    $^{1}$ Columbia University \quad
    $^{2}$ Los Alamos National Laboratory \quad
    $^{3}$ Johns Hopkins University}
}

\iclrfinalcopy %

\begin{document}
\etocdepthtag.toc{main}
\maketitle
\fancyhead{}
\renewcommand{\headrulewidth}{0pt}

\begin{abstract}

Sampling from Boltzmann distributions is a fundamental problem in physics, chemistry, and biology. We introduce \textbf{WA}sserstein \textbf{SA}mpling via \textbf{B}ridge \textbf{I}nterpolants (\methodshort{}), a method that trains a diffusion model to generate samples from a target Boltzmann distribution given access only to the energy function. \methodshort{} minimizes the KL divergence between the distribution generated by its diffusion model and the Boltzmann target using Wasserstein gradient flows. This approach builds on a principle underlying classical sampling methods such as Langevin dynamics. However, applying Wasserstein gradient flows to diffusion models is challenging because the updates require the gradient of the model's own log density, which is generally intractable. We derive a practical training algorithm that avoids computing this gradient. Once trained, \methodshort{} can generate new samples without further evaluations of the energy or its gradient. On alanine dipeptide, \methodshort{} outperforms the best reported baselines using an estimated 3.3 times fewer energy evaluations. It also achieves competitive performance on interacting particle system benchmarks.
\end{abstract}

\section{Introduction}
\label{sec:introduction}

We consider the problem of sampling from a Boltzmann distribution,
\[
\pi(x) = \frac{e^{-U(x)}}{\mathcal Z},
\qquad
\mathcal Z = \int e^{-U(x)}\,dx,
\]
where \(U(x)\) is an energy function and \(\mathcal Z\) is the normalizing constant. We assume no samples from \(\pi\) are available and have only black-box access to \(U(x)\) and its gradient \(\nabla U(x)\).
Many systems in physics, chemistry, and biology are naturally described by Boltzmann distributions. For these systems, samples from $\pi$ are useful because they allow us to estimate expectations $\mathbb{E}_{\pi}[f(X)]$. For appropriate choices of $f$, these expectations answer important scientific questions such as how likely a protein is to fold \citep{noe2009constructing,lindorff2011fast}, how likely a drug is to bind to a particular protein \citep{buch2011high}, or, for a material, what its physical properties are and which of its phases is most stable \citep{frenkel2001understanding,Jarzynski_2002}.

Classical approaches to Boltzmann sampling include Markov chain Monte Carlo  \citep[MCMC;][]{liu2001monte} and molecular dynamics \citep[MD;][]{frenkel2001understanding}. These methods start from an initial state \(x_0\) and repeatedly update it with small, local changes computed from \(U\) or \(\nabla U\), producing a sequence of states \(x_0 \to x_1 \to x_2 \to \cdots\). Langevin dynamics is a canonical example of such a classical approach~\citep{leimkuhler2015molecular}. It updates \(x_t\) with a small step along \(-\nabla U(x_t)\), toward more likely states, plus Gaussian noise. Under suitable conditions, the distribution of \(x_t\) approaches \(\pi\) as \(t \to \infty\).

In practice, however, MCMC methods like Langevin dynamics take a long time to converge to $\pi$. The reason is that moving from one likely region to another may require passing through unlikely states, and such transitions are rare~\citep{tierney1994markov}. Moreover, evaluating \(U\) or \(\nabla U\) at every step can be expensive. In computational chemistry, for example, evaluating the energy of a particular state requires an expensive quantum-mechanical calculation, such as one based on density functional theory \citep{kohn1965self,noe2020machine}.

These costs motivate \emph{amortized samplers}: generative models trained to approximate the Boltzmann distribution. Once trained, these models can produce independent samples cheaply, without evaluating the energy or its gradient. Although training requires such evaluations, their cost is spread, or amortized, across the samples the model later produces. By contrast, obtaining representative samples from $\pi$ with MCMC or MD can be computationally expensive, as both may require long simulations involving repeated energy or gradient evaluations.

In this work, we introduce \textbf{WA}sserstein \textbf{SA}mpling via \textbf{B}ridge \textbf{I}nterpolants
(WASABI), an amortized sampler for Boltzmann distributions. WASABI is trained by minimizing the Kullback--Leibler (KL) divergence
$\mathrm{KL}(q\|\pi)$ between its sampling distribution $q$ and the target
$\pi$. The idea is to perform gradient descent on this divergence directly in the
space of distributions, a procedure known as the Wasserstein gradient flow
\citep[WGF;][]{jko98wgf}. To implement this flow, WASABI represents $q$ with a diffusion model, a
powerful class of generative models that have been successfully applied to
images, video, and protein structures \citep{ho2020denoising,song2021scorebased,ho2022video,watson2023novo}. WASABI repeatedly updates the diffusion model so that its endpoint distribution $q$ takes small steps along the Wasserstein gradient flow toward $\pi$.

\begin{figure}[t]
    \setlength{\abovecaptionskip}{5pt}
    \setlength{\belowcaptionskip}{0pt}
    \centering
    \input{figures/diffusion_figure_denoisers_q0.tex}
    \caption[Overview of \methodshort{}]{Overview of \wasabi{}.
    The endpoint distributions evolve from an initialization $q_0$, distinct from the shared diffusion prior $p_0$.
    \textbf{Top:} we optimize a sequence of sampling distributions $q_n$ by evolve them toward the target $\pi$
    through Wasserstein gradient descent of the KL divergence. \textbf{Bottom:} the sampling distribution $q_n$ is parameterized by a diffusion
    model with a prior $p_0$ for all $n$; \wasabi{} translates Wasserstein descent into
    updates of the diffusion model by updating its denoising representation $\bar m_t^{(n)}$, gradually moving it toward the target across iterations $n$.
    The final panel shows the ideal diffusion sampler, with denoising representation $\bar m_t^\star$ associated with $\pi$.}
    \label{fig:fig1}
\end{figure}

\Cref{fig:fig1} illustrates the idea. WASABI trains a diffusion model to follow Wasserstein gradient flow toward $\pi$. At each step $n$, WASABI seeks to translate a Wasserstein descent step of its current distribution $q_n$ into an update of the diffusion model, yielding a new distribution $q_{n+1}$. Directly implementing this update would require the gradient of the model's log density (the model score), which is generally unavailable for diffusion samplers. We show how to perform this update using only samples drawn from $q_n$ and target energy gradients $\nabla U$, without evaluating the model score. Repeating this procedure yields a diffusion model trained to approximate $\pi$, from which samples can be generated without further energy or gradient evaluations.

We summarize our contributions as follows:
\begin{itemize}
    \item \textbf{Amortizing Langevin dynamics.} We introduce a diffusion sampler that amortizes the evolution of distributions under Langevin dynamics. Formally, this evolution is a Wasserstein gradient flow of the KL divergence to $\pi$ \citep{jko98wgf}. Discretizing this flow gives an update that \methodshort{} applies repeatedly to its distribution; in its exact form, the update has $\pi$ as its fixed point.

    \item \textbf{A tractable training algorithm.} 
    We derive a tractable algorithm for applying  Wasserstein descent-induced updates to the diffusion model that avoids evaluating the diffusion model score. Each step updates the diffusion model through regression, for which we construct unbiased, empirically low-variance estimators of the training targets.

    \item \textbf{Accurate molecular sampling with fewer energy evaluations.}
    On alanine dipeptide, WASABI reproduces the dominant conformational
    basins and achieves lower Jensen--Shannon divergence than the best
    reported baselines \citep{blessing2026bms}, using an estimated
    30\% of their energy/force evaluations.
\end{itemize}

\begingroup
\setlength{\abovedisplayskip}{5pt plus 1pt minus 2pt}
\setlength{\belowdisplayskip}{5pt plus 1pt minus 2pt}
\setlength{\abovedisplayshortskip}{0pt plus 1pt}
\setlength{\belowdisplayshortskip}{4pt plus 1pt minus 1pt}
\section{Background}
\label{sec:background}
\paragraph{Diffusion models and bridge interpolants.}
Diffusion models~\citep{ho2020denoising,song2021scorebased} and related
developments~\citep{lipman2023flow,albergo2025stochastic,heng2025simulating}
provide tools for generative modeling and sampling through continuous-time dynamics.
To sample a target distribution $p_1$, we  simulate the following SDE driven
by a function family $m_t:\mathbb R^d\to\mathbb R^d$:
\begin{equation}\label{eqn:sde-bridge}
dX_t=\frac{m_t(X_t)-X_t}{1-t}\,dt+B\,dW_t,
\qquad X_0\sim p_0,\quad 0\leq t<1,
\end{equation}
where $p_0$ is the source distribution which is easy to sample from, $B>0$ is the diffusion
coefficient, and $W_t$ is standard Brownian motion. The choice of $m_t$ that generates the same endpoint distribution $p_1$ need not be
unique. We consider a \emph{canonical} representation of $p_1$ induced by a \emph{bridge interpolant}:
\begin{equation}\label{eqn:bridge-interpolant}
Y_t=(1-t)\widetilde X_0+t\widetilde X_1+h_tZ,
\qquad 0\leq t\leq1,
\end{equation}
where $\widetilde X_0\sim p_0$, $\widetilde X_1\sim p_1$  are source and endpoint samples and $Z\sim \mathcal N(0,I_d)$ is the noise, all drawn independently;   $h_t=B\sqrt{t(1-t)}$ is the bridge noise standard deviation.
The associated bridge conditional mean defines the  \emph{denoising representation} below.
\begin{definition}[Denoising representation of $p_1$]\label[definition]{def:denoising-representation}
Under the bridge setup in \Cref{eqn:bridge-interpolant}, we call the
family $\bar m_t(y)=\mathbb E[\widetilde X_1\mid Y_t=y]$, $0\leq t<1$,
the \emph{denoising representation} of $p_1$ relative to $p_0$ and $B$.
\end{definition}
Under suitable regularity conditions, choosing $m_t=\bar m_t$ in
\Cref{eqn:sde-bridge} reproduces the bridge marginals
($X_t\overset d=Y_t$) and yields endpoint distribution $p_1$ as $t\uparrow1$. 

When endpoint samples  $\tilde X_1 \sim p_1$ are available, a neural denoiser
$m_\omega(y,t)$ is trained using the denoising regression objective, whose population minimizer is $\bar m_t(y)$:
\begin{equation}
\min_\omega\;
\mathbb E_{t,\widetilde X_0,\widetilde X_1,Z}
\!\left[\|m_\omega(Y_t,t)-\widetilde X_1\|^2\right].
\label{eqn:denoiser-regression}
\end{equation}

\paragraph{Wasserstein gradient flows.}
Let $\mathcal P_2(\mathbb R^d)$ denote the probability measures on $\mathbb R^d$ with finite second moment. In Euclidean space, the gradient describes how a function changes when its input is displaced. Here, we consider a functional $\mathcal F:\mathcal P_2(\mathbb R^d)\to\mathbb R$ that assigns a scalar objective value to a probability distribution, e.g. a divergence to some target distribution. Its gradient likewise
describes how the objective value $\mathcal F$ changes when that distribution is displaced. We displace a distribution $q$ in the direction of a vector field $v$ by moving each point $x$ to $x+\epsilon v(x)$, where $\epsilon>0$ is a small step size, which gives
$q_\epsilon^v =(\mathrm{id}+\epsilon v)_\#q$,
where $\mathrm{id}$ is the identity map and $\#$ denotes the pushforward \citep{santambrogio2015optimal}.\footnote{For a map $T$, the pushforward $T_\#q$ is the probability distribution of $T(X)$ when $X\sim q$.} This notion of displacement allows us to define the \emph{Wasserstein gradient} of $\mathcal F$ at $q$, denoted $\nabla_{W_2}\mathcal F[q]$, as the vector field satisfying
\begin{equation} \label{eqn:w2-gradient}
\left.\frac{d}{d\epsilon}\mathcal F(q_\epsilon^v)\right|_{\epsilon=0}
=
\mathbb E_{X\sim q}\!
\left[\nabla_{W_2}\mathcal F[q](X)\cdot v(X)\right]
\end{equation}
for every sufficiently regular velocity field $v$ \citep{jko98wgf}. 
This identity expresses the directional derivative of
$\mathcal F$ as an $L^2(q)$ inner product; consequently, the negative
Wasserstein gradient gives the direction of steepest descent among velocity
fields $v$ with the same $L^2(q)$-norm $E_q\|v\|^2$.

We say that a sufficiently regular curve $(q_\tau)_{\tau\geq 0}$
follows $2$-Wasserstein gradient flow if, to first order,
\begin{equation}
q_{\tau+\epsilon}
=
\bigl(\mathrm{id}
-\epsilon\nabla_{W_2}\mathcal F[q_\tau]\bigr)_\#q_\tau
+o(\epsilon),
\end{equation}
where the expansion is understood weakly, that is, after integration
against smooth test functions.
Equivalently, when $q_\tau$ admits a density, its evolution
satisfies the continuity equation
\begin{equation}
\partial_\tau q_\tau
=\nabla\cdot\bigl(q_\tau\nabla_{W_2}\mathcal F[q_\tau]\bigr).
\label{eqn:wasserstein-continuity}
\end{equation}

For sampling the Boltzmann target $\pi(x)\propto e^{-U(x)}$, consider the reverse
KL objective $\mathcal F[q]=\mathrm{KL}(q\Vert\pi)$. Its Wasserstein
gradient is
\begin{equation}
\nabla_{W_2}\mathcal F[q](x)
=\nabla\log q(x)-\nabla\log\pi(x)
=\nabla\log q(x)+\nabla U(x).
\label{eqn:kl-wasserstein-gradient}
\end{equation}
Substituting this gradient into \Cref{eqn:wasserstein-continuity} yields the Fokker--Planck equation associated with the following overdamped Langevin dynamics,
\begin{equation}
dZ_\tau=-\nabla U(Z_\tau)\,d\tau+\sqrt{2}\,dW_\tau,
\label{eqn:langevin-dynamics}
\end{equation}
where $W_\tau$ is standard Brownian motion and $\pi$ is the stationary
density. Thus, the distribution of the Langevin process follows the
$2$-Wasserstein gradient flow of the reverse KL \citep{jko98wgf}. 

\paragraph{Stein's identity.}
Let $r$ be a smooth positive density with score $s_r=\nabla\log r$. For any sufficiently regular scalar test function $\xi:\mathbb R^d\to\mathbb R$, Stein's identity gives

\begin{equation}\label{eqn:stein-identity}
\mathbb E_{X\sim r}[\xi(X)s_r(X)]
=-\mathbb E_{X\sim r}[\nabla\xi(X)],
\end{equation}
provided the expectations are finite and the integration-by-parts boundary term vanishes. This identity transfers a derivative of the density to the test function, allowing score-weighted expectations to be expressed without evaluating the score.
\endgroup

\begingroup
\setlength{\abovedisplayskip}{5pt plus 1pt minus 2pt}
\setlength{\belowdisplayskip}{5pt plus 1pt minus 2pt}
\setlength{\abovedisplayshortskip}{0pt plus 1pt}
\setlength{\belowdisplayshortskip}{4pt plus 1pt minus 1pt}
\section{Method}
\label{sec:method}
 
Our goal is to learn a diffusion sampler whose endpoint distribution $q$
approximates the Boltzmann target $\pi\propto e^{-U}$ by minimizing
$\mathcal F[q]=\mathrm{KL}(q\Vert\pi)$.
The key idea is to parameterize the sampling
distribution through a diffusion model's denoising representation and translate the prescribed distributional evolution into an update of this representation. This yields a fixed-point algorithm with regression updates using generated samples from the current model and target energy gradients.

In \Cref{sec:wasserstein-bridge}, we describe the model setup and derive the denoiser update induced by Wasserstein descent. In \Cref{sec:practical-estimators}, we construct estimators of the mean and covariance contributions to the denoiser update that avoid
evaluating the unknown score of the model's distribution. Finally, \Cref{sec:wasabi-practical-algorithm} combines these ingredients into the WASABI training algorithm.

\subsection{Model Setup and Denoiser Updates}
\label{sec:wasserstein-bridge}
We consider the denoising representation $\bar m_t$ of $q$ given a source $p_0$ and a diffusion coefficient $B$ 
(\Cref{def:denoising-representation}). 
WASABI progressively optimizes $\mathrm{KL}(q\Vert\pi)$ by translating Wasserstein
descent steps of the endpoint distribution $q$ into  updates of the denoiser $\bar m_t$.
We index these update iterations by $n=1,2,\cdots$,  writing $\bar m_t^{(n)}$
for the current denoiser and $q_n$ for the endpoint distribution.

 At iteration $n$, the negative Wasserstein gradient (\Cref{eqn:kl-wasserstein-gradient}) specifies the descent direction for $q_n$, with
\begin{align}\label{eq:wgradient-as-scores}
\nabla_{W_2}\mathcal F[q_n]
=\nabla\log q_n - \nabla\log\pi = \nabla \log q_n + \nabla U
\end{align}
For a step size $\eta>0$, the desired endpoint distribution is updated as

\begin{equation}
q_{n,\eta} = \bigl(\mathrm{id}-\eta\nabla_{W_2}\mathcal F[q_n]\bigr)_\#q_n.
\label{eqn:wasserste-descent-q}
\end{equation}

For the same source $p_0$ and diffusion coefficient $B$, the transported distribution $q_{n,\eta}$ induces a new denoising representation $\bar m_{t,\eta}^{(n)}$. In the following theorem, we characterize how the denoiser changes under a small
endpoint transport step by computing its derivative with respect to $\eta$ at $\eta=0$.

\begin{theorem}[Denoiser change induced by Wasserstein descent] \label{thm:denoising-representation}
Let $q_{n,\eta}=(\mathrm{id}-\eta\nabla_{W_2}\mathcal F[q_n])_\#q_n$, and let $(\bar m_{t,\eta}^{(n)})_t$ be its
denoising representation, using the same initial distribution $p_0$ and noise scale $B$, so that $\bar m_{t,0}^{(n)}=\bar m_t^{(n)}$.
Let $h_t=B\sqrt{t(1-t)}$ and $Y_t=(1-t)\widetilde X_0+t\widetilde X_1+h_tZ$ where $\widetilde X_0\sim p_0$, $\widetilde X_1\sim q_n$, and $Z\sim\mathcal N(0,I)$ are independent .  Assume the conditional moments below are finite and differentiation under the conditional integrals is valid. Then, for $0<t<1$,
\begin{equation}
\begin{aligned}
\left.\frac{\partial}{\partial\eta}\bar m_{t,\eta}^{(n)}(y)\right|_{\eta=0}
&=
\underbrace{-\mathbb E\!\left[
\nabla_{W_2}\mathcal F[q_n](\widetilde X_1) \mid Y_t=y\right]}_{\mathcal M_t[q_n](y)}
\\
&\quad+
\underbrace{\frac{t}{h_t}\operatorname{Cov}\!\left(
\widetilde X_1,-Z\cdot\nabla_{W_2}\mathcal F[q_n](\widetilde X_1) \mid Y_t = y
\right)}_{\mathcal C_t[q_n](y)}.
\end{aligned}
\label{eqn:first-order-denoiser-update}
\end{equation}
Moreover, if $q_n=\pi$, then $\nabla_{W_2}\mathcal F[\pi]=0$, and hence $\mathcal M_t[\pi]=\mathcal C_t[\pi]=0$.
\end{theorem}

Transporting the endpoints changes the denoiser 
in two ways. The mean term $\mathcal M_t[q_n](y)$ captures the average movement while keeping each endpoint's original  contribution (under $q_n$-induced process) to the average fixed. But moving an endpoint also
changes how likely the bridge is to produce the same observation $Y_t=y$. Endpoints that become more likely given this
observation contribute more to the conditional average, and those that
become less likely contribute less. The covariance term
$\mathcal C_t[q_n](y)$ accounts for how these changes in the endpoints' relative contributions shift the conditional average. Together, they give the full first-order change
of the denoiser.

\Cref{thm:denoising-representation}  motivates the \emph{WASABI  update operator} for the denoiser. With a time-dependent step size $\eta_t>0$
and a weight $\lambda_t$ controlling the covariance contribution, we define 
\begin{equation}
\mathcal T[\bar m^{(n)}]_t(y)
:=
\bar m_t^{(n)}(y)
+\eta_t\Bigl(
\mathcal M_t[q_n](y)
+\lambda_t\mathcal C_t[q_n](y)
\Bigr).
\label{eqn:wasabi-denoiser-update}
\end{equation}

With $\lambda_t=1$ and a common step size $\eta_t=\eta$, this update
agrees, to first order, with the denoiser induced by the desired transport
in \Cref{eqn:wasserste-descent-q}. Allowing $\eta_t$ and $\lambda_t$ to
vary with bridge time $t$ modifies this direction and its step size. Since both
contributions vanish at $q_n=\pi$, the update leaves the denoiser
of $\pi$ unchanged for any finite $\lambda_t$; hence, $\pi$ is a fixed point of the ideal WASABI update.

\subsection{Estimating the denoiser update} \label{sec:practical-estimators}

The WASABI update direction is now specified
(\Cref{eqn:wasabi-denoiser-update}), but its mean and covariance contributions
(\Cref{eqn:first-order-denoiser-update}) involve the unknown endpoint score
$\nabla\log q_n$ through the Wasserstein gradient (\Cref{eq:wgradient-as-scores}).
We derive score-free estimators for both contributions, assuming the bridge's source and
endpoint are drawn independently. \Cref{app:mean-stein,app:covariance-general-coupling} extend these estimators to general couplings.

\paragraph{Mean contribution.}
The first component of the denoiser update in \Cref{eqn:wasabi-denoiser-update} can be written as \begin{equation}
\mathcal M_t[q_n](y)
=
\mathbb E\!\left[
-\nabla U(\widetilde X_1)
-\nabla\log q_n(\widetilde X_1)
\,\middle|\,Y_t=y
\right],
\label{eqn:mean-update-score}
\end{equation}
 where we substitute the Wasserstein gradient from \Cref{eq:wgradient-as-scores}.

The target score $-\nabla U$ is available through energy
gradients.
Using Stein's identity (\Cref{eqn:stein-identity}), we obtain the
following score-free expression (\Cref{app:mean-stein} gives the derivation):
\begin{equation}
\mathcal M_t[q_n](y)
=
\mathbb E\!\left[
-\nabla U(\widetilde X_1)+\frac{t}{h_t}Z
\,\middle|\,Y_t=y
\right].
\label{eqn:mean-estimator-score-free}
\end{equation}

This expression provides a computable regression target:
draw independent $\widetilde X_0\sim p_0$, $\widetilde X_1\sim q_n$, and
$Z\sim\mathcal N(0,I_d)$, construct
$Y_t=(1-t)\widetilde X_0+t\widetilde X_1+h_tZ$, and compute
$-\nabla U(\widetilde X_1)+(t/h_t)Z$.
Regressing this target on $(Y_t,t)$ with squared-error loss learns its
conditional mean, $\mathcal M_t[q_n](Y_t)$, without evaluating the unknown
endpoint score $\nabla \log q_n(\tilde X_1)$.

\paragraph{Covariance contribution.}
To learn the covariance contribution $\mathcal C_t[q_n](y)$ in \Cref{eqn:first-order-denoiser-update} by regression, we need not evaluate
it pointwise. In \Cref{sec:wasabi-practical-algorithm} we show that expanding the regression loss requires
only the expected inner product between $\mathcal C_t[q_n](y)$ and a learned correction network $f_\theta$. 
To estimate this inner product, here we derive an estimator for a general
smooth vector test field $\varphi:\mathbb R^d\to\mathbb R^d$; in \Cref{sec:wasabi-practical-algorithm}  we take $\varphi=f_\theta(\cdot,t)$.
Write $s_\pi=-\nabla U$ and define
$F_\varphi(y,x):=\varphi(y)\cdot(x-\bar m_t^{(n)}(y))$,
the scalar projection of the centered endpoint onto $\varphi(y)$.
Stein's identity yields the following score-free expression:

\begin{equation}
\mathbb E\!\left[\varphi(Y_t)\cdot\mathcal C_t[q_n](Y_t)\right]
=\frac{t}{h_t}\mathbb E\!\left[
\varphi(Y_t)\cdot Z+
F_\varphi(Y_t,\widetilde X_1)
\left(Z\cdot s_\pi(\widetilde X_1)
+\frac{t}{h_t}(\|Z\|^2-d)\right)
\right].
\label{eqn:covariance-direct-test}
\end{equation}
The derivation is provided in \Cref{apx:cov_expansion}.

Although this expression removes the unknown score, direct Monte Carlo
estimates had high variance in our early experiments. We seek an
alternative representation of the same expectation by changing how
bridge noise is sampled. Completing the square, with
$a_t(x):=\frac{h_t}{2t}s_\pi(x)$, gives
$Z\cdot s_\pi(x)+\frac{t}{h_t}(\|Z\|^2-d)
=\frac{t}{h_t}(\|Z+a_t(x)\|^2-d-\|a_t(x)\|^2)$.

Completing the square isolates the nonnegative factor
$\|Z+a_t(x)\|^2$. We use this factor to reweight the standard Gaussian
noise distribution: noise values with larger factors are sampled more often.
Normalizing these weights gives the tilted distribution in the following theorem.

\begin{theorem}[Tilted-noise covariance estimator]
\label{thm:covariance-tilted-main}

Fix $0<t<1$ and independent bridge endpoints
$\widetilde X_0\sim p_0$, $\widetilde X_1\sim q_n$.
Let $\phi$ be the standard Gaussian density on $\mathbb R^d$, and let
$\varphi:\mathbb R^d\to\mathbb R^d$ be a smooth test field.
Write $a_t(x)=h_ts_\pi(x)/(2t)$ and
$F_\varphi(y,x)=\varphi(y)\cdot(x-\bar m_t^{(n)}(y))$. Then
\begin{equation}
\widetilde\phi_t(z\mid x)
=\frac{\|z+a_t(x)\|^2}{d+\|a_t(x)\|^2}\,\phi(z)
\label{eqn:covariance-tilted-density-main}
\end{equation}
is a probability density.
 Conditional on $(\widetilde X_0,\widetilde X_1)$,
let $Z$ and $\widetilde Z$ have respective distributions $\phi$ and
$\widetilde\phi_t(\cdot\mid\widetilde X_1)$, with any coupling.
Set $\widetilde Y_t=(1-t)\widetilde X_0+t\widetilde X_1+h_t\widetilde Z$.
The estimator
\begin{equation}
\widehat{\mathcal C}_t[\varphi]
=\frac{t}{h_t}\varphi(Y_t)\cdot Z
+\frac{t^2}{h_t^2}
\bigl(d+\|a_t(\widetilde X_1)\|^2\bigr)
\bigl[F_\varphi(\widetilde Y_t,\widetilde X_1)
-F_\varphi(Y_t,\widetilde X_1)\bigr]
\label{eqn:covariance-tilted-estimator-main}
\end{equation}
is an unbiased estimator of
$\mathbb E[\varphi(Y_t)\cdot\mathcal C_t[q_n](Y_t)]$,
provided the relevant expectations are finite, the integration-by-parts
boundary terms vanish, and
$\bar m_t^{(n)}(y)=\mathbb E[\widetilde X_1\mid Y_t=y]$.
\end{theorem}

The proof is given in \Cref{apx:cov_expansion};
\Cref{lem:tilted-sampling} provides a procedure for sampling
the tilted noise.

The tilted distribution (\Cref{eqn:covariance-tilted-density-main})
replaces weighting ordinary Gaussian samples by sampling noise values in
proportion to their weights and multiplying by the known normalization
constant $d+\|a_t(x)\|^2$.
The estimator compares evaluations under the original and tilted noise
distributions. The two noise draws may be coupled so that the estimator uses paired evaluations at $Y_t$ and $\widetilde Y_t$. Its expectation is unchanged by the coupling, but its variance can depend on that choice.

\subsection{The WASABI Algorithm}
\label{sec:wasabi-practical-algorithm}

We now describe the WASABI algorithm, which realizes Wasserstein descent
through learned denoiser updates and canonicalizes the updated sampler by
refitting its denoising representation.

At a high level, WASABI starts from an endpoint sample set $\mathcal D_0$
drawn from an initial distribution $q_0$; in practice, it can be obtained from an uninformative Gaussian or a high temperature MD simulation.  We use $\mathcal D_0$ to fit an 
initial denoiser $m_{\omega_0}$ via \Cref{eqn:denoiser-regression}.
At iteration $n$, we learn the direction of the WASABI denoiser update in \Cref{eqn:wasabi-denoiser-update} by training a correction network $f_\theta$ while keeping the current denoiser $m_{\omega_n}$ fixed. Applying the learned
correction gives the new distribution $q_{n+1}$ by sampling from the diffusion SDE (\Cref{eqn:sde-bridge}) 
with
$m_t(y)=\widetilde m_t^{(n+1)}(y):=m_{\omega_n}(y,t)+\eta_t f_{\theta_n}(y,t)$.
This corrected function $\widetilde m^{(n+1)}$ need not be the denoising
representation of $q_{n+1}$. We therefore \emph{canonicalize} it by refitting
the denoiser $m_{\omega_{n+1}}$ to new samples from $q_{n+1}$, preparing it for the next iteration.
After $N$ outer iterations, we return the diffusion sampler driven by $m_{\omega_N}$.

\Cref{alg:wasabi-main} summarizes the procedure.
We derive the training objectives and practical update rules below.

\paragraph{Learning the Wasserstein-induced update.}
Below we state the population objectives for the ideal representation
$\bar m^{(n)}$; implementation uses its fitted approximation $m_{\omega_n}$.

To approximate the first-order denoiser change in
\Cref{thm:denoising-representation}, we train the correction network to fit
the update direction prescribed by \Cref{eqn:wasabi-denoiser-update}:

\begin{equation}
r_t^{(n)}(y)
:=\eta_t^{-1}\left(\mathcal T[\bar m^{(n)}]_t(y)-\bar m_t^{(n)}(y)\right)
=\mathcal M_t[q_n](y)+\lambda_t\mathcal C_t[q_n](y).
\label{eqn:wasabi-correction-direction}
\end{equation}

Fitting this direction amounts to the ideal regression problem

\begin{equation}
\min_\theta\;\frac12\mathbb E\!\left[
\left\|f_\theta(Y_t,t)-r_t^{(n)}(Y_t)\right\|^2
\right].
\label{eqn:wasabi-ideal-regression}
\end{equation}
Here the expectation is over the sampled bridge time $t$ and independent
$\widetilde X_0\sim p_0$, $\widetilde X_1\sim q_n$, and
$Z\sim\mathcal N(0,I_d)$, with $Y_t$ constructed as in the background.

We do not need to evaluate $r_t^{(n)}(Y_t)$ itself to train the correction.
Expanding the squared-error objective in \Cref{eqn:wasabi-ideal-regression}
and dropping the term independent of $\theta$, we replace its mean and
covariance inner-product terms using the estimators in
\Cref{eqn:mean-estimator-score-free,eqn:covariance-tilted-estimator-main}.
This gives the following computable objective with the same population
minimizers (\Cref{app:regression-objective} provides the derivation): 

\begin{equation}
\mathcal L_n(\theta)
=\mathbb E\!\left[
\frac12\|f_\theta(Y_t,t)\|^2
-f_\theta(Y_t,t)\cdot
\left(-\nabla U(\widetilde X_1)+\frac{t}{h_t}Z\right)
-\lambda_t\widehat{\mathcal C}_t[f_\theta(\cdot,t)]
\right].
\label{eqn:wasabi-practical-loss}
\end{equation}

This expectation additionally includes the tilted noise $\widetilde Z$,
whose conditional distribution given the source and endpoint samples is
specified in \Cref{thm:covariance-tilted-main}.

We minimize this objective with minibatch estimates, keeping
$m_{\omega_n}$ fixed inside the covariance estimator.
With the fitted parameters $\theta_n$, we form
$\widetilde m_t^{(n+1)}(y)=m_{\omega_n}(y,t)+\eta_t f_{\theta_n}(y,t)$.

\paragraph{Refitting the denoising representation.}
With $\widetilde m^{(n+1)}$, the corrected sampler  generates
endpoint distribution $q_{n+1}$ via \Cref{eqn:sde-bridge}, but $\widetilde m^{(n+1)}$ needs not equal
the denoising representation of $q_{n+1}$, which is required for the next update.
We therefore canonicalize the corrected sampler by refitting the denoiser.
To do so, we simulate samples $\mathcal D_{n+1}$ from $q_{n+1}$, then fit $m_{\omega_{n+1}}$ using the
denoising regression objective in \Cref{eqn:denoiser-regression}.

\begin{algorithm}[t]
\normalcolor
\caption{WASABI}
\label{alg:wasabi-main}
\algrenewcommand{\algorithmicrequire}{\textbf{Inputs:}}
\begin{algorithmic}[1]
\Require Energy $U$; %
 source $p_0$; diffusion coefficient $B$; %
 initial endpoint sample set $\mathcal D_0$;
iterations $N$; schedules $\eta_t,\lambda_t$.
\State \textbf{Fit initial denoiser:} fit $m_{\omega_0}$ to
$\mathcal D_0$ using \Cref{eqn:denoiser-regression}.
\For{$n=0,\ldots,N-1$}
    \Statex \textit{// Wasserstein-induced update}
    \State \textbf{Fit correction:} freeze $m_{\omega_n}$ and minimize
    $\mathcal L_n(\theta)$ in \Cref{eqn:wasabi-practical-loss} to obtain $\theta_n$.
    \State \textbf{Update sampler:}
    $\widetilde m_t^{(n+1)}(y)\gets m_{\omega_n}(y,t)+\eta_t f_{\theta_n}(y,t)$.
    \Statex \textit{// Canonicalization}
    \State \textbf{Refresh sample set:} generate $\mathcal D_{n+1}$
    by simulating \Cref{eqn:sde-bridge} with $\widetilde m^{(n+1)}$.
    \State \textbf{Fit denoiser:}
    $\omega_{n+1}\approx\arg\min_\omega\mathbb E[\|m_\omega(Y_t,t)-\widetilde X_1\|^2]$,
    using $\widetilde X_1\sim\mathcal D_{n+1}$.
\EndFor
\State \Return Diffusion sampler driven by $m_{\omega_N}$.
\end{algorithmic}
\end{algorithm}

\paragraph{Replay and annealing.}
In practice, we maintain the endpoint sample set as a replay buffer,
retaining samples across iterations and partially refreshing it with
samples from the updated sampler.
Batches from this buffer are used to fit both the denoiser and its
correction. Because retained samples may come from earlier samplers,
this reuse approximates the population iteration written in terms of
$q_n$. For difficult targets, we also use annealing:
$U_{\beta_n}=\beta_n U$, with $\beta_n$ increasing toward $1$.
At each stage, the target score $-\beta_n\nabla U$ replaces $-\nabla U$
in both estimators, allowing the sampler to adapt through a sequence
of tempered targets. \Cref{alg:wasabi-extended} gives the extended
procedure with replay, optional alignment, and annealing.

\endgroup

\section{Related work}
\label{sec:related_work}

We focus on diffusion-based methods that learn amortized samplers for unnormalized targets using access to the target energy and/or its gradients, without requiring a pre-generated target dataset. PIS \citep{zhang2022path} formulates sampling as stochastic optimal control and trains a controlled diffusion with a path-space objective. DDS \citep{vargas2023denoising} constructs a noising process from the target to a tractable reference and learns the
corresponding time reversal. iDEM \citep{pmlr-v235-akhound-sadegh24a} estimates scores of noised target
distributions from energy evaluations and iteratively refines a diffusion model using its own samples.

More recent methods use bridge-based regression to iteratively refine the sampler. Adjoint Sampling \citep[AS;][]{havens2025as} uses reciprocal adjoint matching to regress diffusion drifts while reusing endpoint energy gradients across
reference bridges. Adjoint Schr\"odinger Bridge Sampler \citep[ASBS;][]{liu2025asbs} extends this construction to general source distributions through a Schrodinger bridge formulation with a learned terminal corrector. Bridge Matching Sampler  \citep[BMS;][]{blessing2026bms} introduces
a generalized fixed-point matching framework that recovers AS and ASBS, with an independent-coupling construction yielding explicit regression targets without a separate corrector. Flow Sampling \citep{havens2026flow} similarly learns denoising drifts from source-conditioned probability paths using regression targets constructed from generated endpoints and endpoint energy gradients. WASABI likewise learns through regression using generated samples and target energy gradients. However, whereas the preceding methods derive regression targets from stochastic-control optimality or bridge-based matching identities, WASABI derives its updates from the  Wasserstein gradient flow.

Finally, several works, like WASABI, train diffusion samplers through a prescribed evolution of the sampling distribution. PITA \citep{akhound2025pita} progressively trains a sequence of diffusion models along a temperature-annealing path from high temperature to the target temperature, using Sequential Monte Carlo (SMC)-based inference-time annealing to generate samples for successive model refitting. WASABI instead learns Wasserstein-induced updates through regression without SMC reweighting. Tilt Matching similarly derives regression updates to a vector-field representation of the model, but follows an exponentially tilted path induced by a reward rather than WASABI’s Wasserstein descent of the KL divergence to the target.  Closely related concurrent work, Newton Matching \citep{li2026newton}, optimizes the same KL objective under Fisher--Rao geometry and focuses on theory and algorithms without empirical evaluation. In comparison, WASABI uses Wasserstein geometry and is evaluated on sampling benchmarks involving particle systems and alanine dipeptide.

\section{Experiments}
\label{sec:experiments}

Following previous literature, we evaluate WASABI on three benchmark Boltzmann distributions: the four-particle double well (DW-4), a 13-particle Lennard–Jones cluster (LJ-13), and alanine dipeptide (ALA2) \citep{kohler2020equiflows, midgley2023flow}. We assess how well it reproduces each target’s energy and structure, and compare its performance to existing diffusion samplers.

Our comparison methods are Adjoint Sampling \citep[AS;][]{havens2025as}, Adjoint Schr\"odinger Bridge Sampler \citep[ASBS;][]{liu2025asbs}, and Bridge Matching Sampler \citep[BMS;][]{blessing2026bms}, including damped variants where reported. For LJ-13 and ALA2, baseline results are taken from \citet{blessing2026bms}; the DW-4 methods are trained under matched settings. Experiment details are given in Appendix~\ref{app:experiment-details}.

\subsection{Double-well system (DW-4)}

\begingroup
\setlength{\intextsep}{4pt}
\begin{figure}[!htbp]
\setlength{\abovecaptionskip}{5pt}
\setlength{\belowcaptionskip}{0pt}
\centering
\begingroup
\definecolor{dwresultgreen}{HTML}{347A46}
\noindent\begin{minipage}[c]{0.43\linewidth}
  \includegraphics[width=\linewidth]{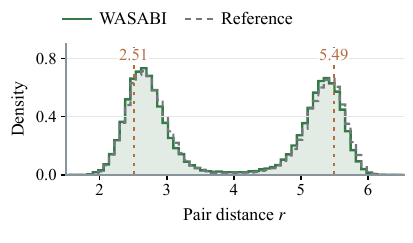}
\end{minipage}\hfill
\begin{minipage}[c]{0.55\linewidth}
  \refstepcounter{table}\label{tab:dw4-preliminary}
  {\normalsize\textbf{Table~\thetable: DW-4 at 1,000 rounds.}\par}
  {\normalsize $16\times4$ EGNN; matched training settings.\par}
  \vspace{11pt}
  {\normalsize
  \setlength{\tabcolsep}{2pt}
  \renewcommand{\arraystretch}{1.08}
  \begin{tabular*}{\linewidth}{@{\extracolsep{\fill}}lcc@{}}
  \toprule
  Method & Config. $W_2\downarrow$ & Energy $W_2\downarrow$\\
  \midrule
  ASBS & $\mathbf{0.94\pm0.16}$ & $2.35\pm0.73$\\
  ASBS (damped) & $0.97\pm0.30$ & $2.31\pm0.34$\\
  BMS & $1.27\pm0.19$ & $2.87\pm1.23$\\
  BMS (damped)$^{\dagger}$ & $1.48\pm0.21$ & $2.72\pm0.58$\\
  \midrule
  \textcolor{dwresultgreen}{\textbf{WASABI}} & $1.13\pm0.39$ &
    \textcolor{dwresultgreen}{$\mathbf{1.89\pm0.43}$}\\
  \bottomrule
  \end{tabular*}}
  \vspace{2pt}
  {\normalsize Mean $\pm$ SD over six seeds.
  $^{\dagger}$Five seeds; one divergent run excluded
  (energy $W_2=2{,}449$).\par}
\end{minipage}
\endgroup
\caption{\textbf{DW-4. (Left)} Pooled pairwise distances $\{r_{ij}\}_{i<j}$
from 10,000 reference and 10,000 \textcolor[HTML]{347A46}{WASABI} samples;
dashed lines mark the pair-potential minima. \textbf{(Right)} Matched-training
comparison using samples pooled over rounds 625--1,000. $W_2$ compares
particle configurations (Config.) and unfiltered potential energies (Energy).}
\label{fig:dw4-pair-distances}
\end{figure}
\endgroup

We first consider DW-4, a two-dimensional system of four particles with
pairwise double-well interactions (Appendix~\ref{apx:particle-potentials}).
Each pairwise term is minimized at two distances, $r_{ij}\approx2.51$ and
$5.49$, favoring competing structural arrangements. Interactions among all
four particles can shift the peaks of the pair-distance distribution,
although we observe them near these minima.
WASABI reproduces both peaks of the pair-distance distribution
(\Cref{fig:dw4-pair-distances}). Under matched training settings, it achieves
the lowest mean energy $W_2$, while ASBS achieves the lowest configuration
$W_2$, indicating that the methods differ in how accurately they reproduce
energies and particle arrangements (\Cref{tab:dw4-preliminary}).

\subsection{Lennard--Jones cluster (LJ-13)}

\begingroup
\setlength{\intextsep}{4pt}
\begin{figure}[!htbp]
\setlength{\abovecaptionskip}{5pt}
\setlength{\belowcaptionskip}{0pt}
\centering
\begingroup
\definecolor{ljresultgreen}{HTML}{347A46}
\noindent\begin{minipage}[c]{0.48\linewidth}
  \includegraphics[width=\linewidth]{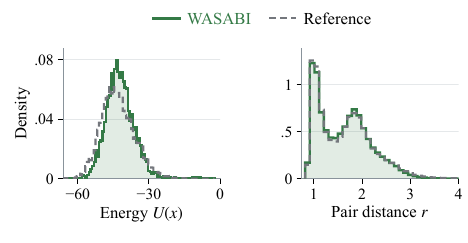}
\end{minipage}\hfill
\begin{minipage}[c]{0.49\linewidth}
  \refstepcounter{table}\label{tab:lj13-comparison}
  {\normalsize\textbf{Table~\thetable: LJ-13 comparison.}\par}
  \vspace{11pt}
  {\normalsize
  \setlength{\tabcolsep}{2pt}
  \renewcommand{\arraystretch}{1.00}
  \setlength{\aboverulesep}{1.5pt}
  \setlength{\belowrulesep}{1.5pt}
  \begin{tabular*}{\linewidth}{@{\extracolsep{\fill}}lcc@{}}
  \toprule
  Method & Config. $W_2\downarrow$ & Energy $W_2\downarrow$\\
  \midrule
  AS & $1.67\pm0.01$ & $2.40\pm1.25$\\
  ASBS & $1.59\pm0.03$ & $1.99\pm1.01$\\
  BMS & $\mathbf{1.54\pm0.00}$ & $\mathbf{0.53\pm0.04}$\\
  \midrule
  \textcolor{ljresultgreen}{\textbf{WASABI}} &
  \textcolor{ljresultgreen}{$1.597\pm0.003$} &
  \textcolor{ljresultgreen}{$1.677\pm0.224$}\\
  \bottomrule
  \end{tabular*}}
\end{minipage}
\endgroup
\caption{\textbf{LJ-13. (Left)} Energy and pooled pair-distance
$\{r_{ij}\}_{i<j}$ densities: \textcolor[HTML]{347A46}{WASABI}
(green, three-seed mean) and reference (dashed).
\textbf{(Right)} Configuration $W_2$ compares particle arrangements after
matching and alignment; energy $W_2$ compares energy distributions.
Results are mean $\pm$ SD across three seeds; baselines are quoted from
\citet[Table~2]{blessing2026bms}.}
\label{fig:lj13-results}
\end{figure}
\endgroup

We next consider 13 identical particles in three dimensions with
Lennard--Jones interactions and quadratic confinement
(Appendix~\ref{apx:particle-potentials}).
The pair interaction strongly repels nearby particles and favors a separation of \(r_{ij}=1\), while the quadratic term penalizes dispersed clusters.

We train a center and a correction PaiNN, each with four layers and 64
channels, from a Gaussian prior with scale $\sigma_0=1.75$. We evaluate
three seeds after 7,000 outer iterations, using 128 sampling steps.
WASABI closely matches the pooled pair-distance distribution, although its
energy density is more concentrated near the peak and underrepresents the
low-energy shoulder (\Cref{fig:lj13-results}). Its configuration $W_2$ is
close to ASBS, and its mean energy $W_2$ is lower than AS and ASBS, while BMS
performs better on both metrics (\Cref{tab:lj13-comparison}).

\subsection{Alanine dipeptide (ALA2)}
\label{sec:ala2-experiments}

\begingroup
\setlength{\intextsep}{4pt}
\begin{figure}[!htbp]
\setlength{\abovecaptionskip}{5pt}
\setlength{\belowcaptionskip}{0pt}
\centering
\begingroup
\definecolor{alaresultgreen}{HTML}{347A46}
\noindent\begin{minipage}[t]{0.38\linewidth}
  \vspace{0pt}
  \includegraphics[width=\linewidth]{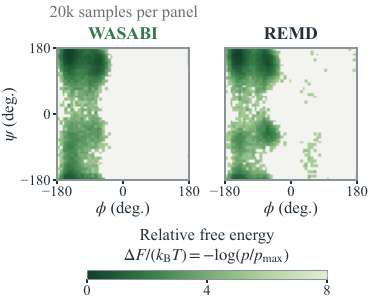}
\end{minipage}\hfill%
\begin{minipage}[t]{0.60\linewidth}
  \vspace{0pt}
  \refstepcounter{table}\label{tab:ala2-comparison}
  {\normalsize\textbf{Table~\thetable: Accuracy and computational cost.}\par}
  \vspace{2pt}
  {\normalsize
  \setlength{\tabcolsep}{1.5pt}
  \renewcommand{\arraystretch}{1.10}
  \begin{tabular*}{\linewidth}{@{\extracolsep{\fill}}lrrr@{}}
  \toprule
  Method & Train $\downarrow$ & Energy $\downarrow$ & $D_{\rm JS}\downarrow$\\
  \midrule
  ASBS & --- & $10.24$\textsuperscript{\ref{fn:ala2-baseline-cost}} & $0.232\pm0.018$\\
  ASBS (damped) & --- & $10.24$\textsuperscript{\ref{fn:ala2-baseline-cost}} & $0.111\pm0.017$\\
  BMS & $112.9$\textsuperscript{\ref{fn:ala2-baseline-cost}} & $10.24$\textsuperscript{\ref{fn:ala2-baseline-cost}} & $0.311\pm0.050$\\
  BMS (damped) & $190.2$\textsuperscript{\ref{fn:ala2-baseline-cost}} & $10.24$\textsuperscript{\ref{fn:ala2-baseline-cost}} & $0.068\pm0.009$\\
  \midrule
  \textcolor{alaresultgreen}{\textbf{WASABI}} & \textbf{31.7} & \textbf{3.13}\textsuperscript{\ref{fn:ala2-wasabi-cost}} &
    \textcolor{alaresultgreen}{\textbf{0.0254\,$\pm$\,0.0033}}\\
  \midrule
  REMD & --- & $73.5$ & ---\\
  \bottomrule
  \end{tabular*}}
  \par\smallskip
  {\footnotesize\raggedright
  \refstepcounter{tablenote}\label{fn:ala2-baseline-cost}%
  \textsuperscript{\thetablenote}Inferred eight-GPU setting based on the
  \href{https://github.com/DenisBless/BridgeMatchingSampler}{released codebase}.\par
  \refstepcounter{tablenote}\label{fn:ala2-wasabi-cost}%
  \textsuperscript{\thetablenote}3.13M calls: 2M prior production + 0.1M equilibration +
  1.03M training (4,000 outer steps; bank of 4,096; 256 fresh/step).\par}
\end{minipage}
\endgroup
\caption{\textbf{ALA2 at 300 K. (Left)} Ramachandran plots comparing
20k \textcolor[HTML]{347A46}{WASABI} and 20k REMD samples; gray bins are
unobserved. \textbf{(Right)} Jensen--Shannon divergence and estimated
computational cost. WASABI results are mean $\pm$ SD across three seeds;
baselines are quoted from \citet{blessing2026bms}. Training cost is in TFLOPs per outer iteration, energy/force calls in millions, and $D_{\rm JS}$ in nats; --- denotes unreported values.}
\label{fig:ala2-results}
\end{figure}
\endgroup

We finally evaluate WASABI on alanine dipeptide (ALA2). Its energy function
includes terms describing bond stretching and angle bending, as well as
preferred molecular conformations, electrostatic interactions, and the
surrounding solvent (details in Appendix~\ref{apx:ala2-hamiltonian}).
As in prior work~\citep{blessing2026bms}, we compare the sampled energy and
backbone-angle distributions with molecular-dynamics reference samples.

Throughout our ALA2 experiments, we use molecular-dynamics configurations
sampled at 1,000\,K as the prior. During training, each sampled prior
configuration is Kabsch-aligned to its paired endpoint from the replay
buffer.
The resulting coupling and corresponding mean and covariance
estimators are derived in
Appendices~\ref{app:kabsch-coupling}--\ref{app:covariance-general-coupling}.
Generating these samples on an Apple M5 Max
chip takes approximately five minutes using one CPU thread. For the main results and
hyperparameter sweeps, we train for 4,000 outer iterations, annealing the
target from 1,000\,K to 300\,K over the first 1,500 iterations. We maintain
a ring replay buffer of 4,096 endpoints and replace 256 endpoints with
fresh samples at each outer iteration.

Using these settings, WASABI achieves a Jensen--Shannon divergence of
$0.0254\pm0.0033$, using the same evaluation procedure as ASBS and
BMS~\citep{liu2025asbs,blessing2026bms}, with one million generated samples
per seed and one million reference configurations.
WASABI trains for 4,000 outer iterations at an estimated 31.7 TFLOPs per
iteration. BMS reports 5,000 outer iterations for ALA2~\citep{blessing2026bms};
under the inferred eight-GPU configuration, we estimate costs of 112.9 and
190.2 TFLOPs per iteration for its undamped and damped variants,
respectively (Table~\ref{tab:ala2-comparison}).
Figure~\ref{fig:ala2-results} and Table~\ref{tab:ala2-comparison} summarize
our results.

\paragraph{Influence of hyperparameters.}
We examine the effects of covariance weight, update step size, and
replay-buffer settings in Appendix~\ref{apx:ala2-results}.
Among the settings tested, WASABI performs best on ALA2 with covariance
weights below $0.5$. To limit the computational cost of the covariance-weight
and step-size ablations, which comprised 96 training runs requiring
approximately 8--10 GPU-hours each, we used a replay-buffer capacity of
4,096 endpoints and refreshed 256 endpoints per outer iteration.
Low Jensen--Shannon divergence can coexist with occasional
underrepresentation of a smaller backbone-angle mode. Increasing
buffer capacity to 8,192 and the refresh count to 512 mitigates this
behavior in the evaluated runs.

\section{Conclusion}
\label{sec:discussion}

In this paper, we introduced WASABI, a method for training diffusion samplers for Boltzmann distributions using Wasserstein gradient flow. We derived a practical training algorithm that avoids computing the diffusion model's intractable log-density gradient. On alanine dipeptide, WASABI achieves lower Jensen--Shannon divergence than the best reported ASBS/BMS variant while using an estimated 3.3 times fewer energy/force evaluations. 
Under matched training settings on DW-4, WASABI also achieves the lowest mean energy $W_2$. 

Future work can investigate scaling WASABI to larger molecular systems, where complex energy landscapes make accurate sampling particularly challenging. Another promising direction is to extend WASABI to transferable sampling, where a single model learns to sample from a family of Boltzmann distributions rather than a fixed target distribution. This setting arises naturally when the potential energy \(U(x;c)\) depends on additional parameters \(c\), such as chemical composition \citep{cheng2026atlas} or molecular identity \citep{klein2024transferable}. By conditioning the diffusion sampler on these parameters, one could train a model that generalizes to previously unseen conditions at test time without retraining. 

More broadly, WASABI shows how the distributional evolution underlying classical sampling methods can be translated into a practical learning rule for diffusion models. Beyond Boltzmann sampling, one promising direction is to extend this approach to fine-tuning pretrained diffusion models toward reward-tilted distributions.

\paragraph{Limitations}
WASABI uses first-order denoiser updates, and finite model fitting and replay-buffer reuse introduce further approximations to the ideal population iteration. On LJ-13, WASABI underrepresents the low-energy shoulder and trails BMS on both configuration and energy Wasserstein distances. On alanine dipeptide, low Jensen--Shannon divergence can coexist with underrepresentation of a smaller conformational mode; increasing replay capacity and refresh count mitigates this behavior in the evaluated runs but does not ensure complete mode coverage.

\clearpage

\section*{Acknowledgements}
DF was partially supported by Soledad Villar’s NSF CAREER award (NSF CAREER 2339682). DB was supported by NSF grants IIS-2127869 and DMS-2311108, Office of Naval Research (ONR) grant N000142412243, and the Simons Foundation. This work was conducted under the U.S. Department of Energy National Nuclear Security Administration’s Office of Defense Nuclear Nonproliferation Research and Development (NA-22) Steel Thread Venture. This research also utilized resources provided by the Los Alamos National Laboratory Institutional Computing Program, which is supported by the U.S. Department of Energy National Nuclear Security Administration under Contract No. 89233218CNA000001. We are deeply grateful to the members of the Blei Lab for their invaluable feedback on early drafts of this manuscript.

\bibliography{iclr2026_conference}
\bibliographystyle{iclr2027_conference}

\clearpage
\appendix
\etocdepthtag.toc{appendix}
\etocsettagdepth{main}{none}
\etocsettagdepth{appendix}{subsection}
\etocsettocstyle{\subsection*{Appendix Contents}}{}
\tableofcontents

\section{Method}
\label{sec:Appendix}
\label{app:method}
\subsection{Denoiser Response to Endpoint Transport}
\label{app:denoiser-update}

\begin{proposition}[Denoiser response to endpoint transport]
Let $(X_0,X_1)\sim\gamma_n$ have marginals $p_0,q_n$, and let
$v:\mathbb R^d\to\mathbb R^d$ be measurable, with
$\mathbb E[\|X_0\|^2+\|X_1\|^2+\|v(X_1)\|^2]<\infty$.
Define $X_1^\epsilon=X_1+\epsilon v(X_1)$, whose distribution is
$q_{n,\epsilon}=(\mathrm{id}+\epsilon v)_\#q_n$.

For fixed $t\in(0,1)$, let $h_t=B\sqrt{t(1-t)}$, $B>0$, and let
$Z\sim\mathcal N(0,I_d)$ be independent of $(X_0,X_1)$.
Define $Y_t^\epsilon=(1-t)X_0+tX_1^\epsilon+h_tZ$ and
$\bar m_{t,\epsilon}^{(n)}(y)=\mathbb E[X_1^\epsilon\mid Y_t^\epsilon=y]$.
Then,
\begin{equation}
\left.\partial_\epsilon \bar m_{t,\epsilon}^{(n)}(y)\right|_{\epsilon=0}
=
\mathbb E[v(X_1)\mid Y_t=y]
+\frac{t}{h_t}\operatorname{Cov}\!\left(
X_1,Z\cdot v(X_1)\mid Y_t=y
\right),
\label{eq:denoiser-transport-variation}
\end{equation}
where $Y_t=Y_t^0$.
\end{proposition}

\begin{proof}
Fix $t\in(0,1)$ and $y\in\mathbb R^d$. Conditional on $X_0=x_0$ and
$X_1=x_1$, the perturbed interpolant has Gaussian density,
\begin{equation*}
K_\epsilon(y\mid x_0,x_1)
=
\frac{1}{(2\pi h_t^2)^{d/2}}
\exp\!\left(
-\frac{\|y-(1-t)x_0-t[x_1+\epsilon v(x_1)]\|^2}{2h_t^2}
\right).
\end{equation*}
Consequently,
\begin{align*}
\bar m_{t,\epsilon}^{(n)}(y)
&=\mathbb E[X_1^\epsilon\mid Y_t^\epsilon=y]\\
&=
\frac{\displaystyle\int
[x_1+\epsilon v(x_1)]K_\epsilon(y\mid x_0,x_1)\,
d\gamma_n(x_0,x_1)}
{\displaystyle\int
K_\epsilon(y\mid x_0,x_1)\,d\gamma_n(x_0,x_1)}
=:\frac{N(\epsilon)}{D(\epsilon)}.
\end{align*}
The denominator is strictly positive because the Gaussian kernel is positive.

To differentiate under these integrals, we establish integrable bounds
for their derivatives. Differentiating the Gaussian gives
\begin{equation*}
\partial_\epsilon K_\epsilon
=
\frac{t}{h_t^2}
\bigl[y-(1-t)x_0-t(x_1+\epsilon v(x_1))\bigr]\cdot v(x_1)\,
K_\epsilon.
\end{equation*}
By the product rule,
\begin{align*}
\partial_\epsilon\!\left([x_1+\epsilon v(x_1)]K_\epsilon\right)
&=v(x_1)K_\epsilon
+[x_1+\epsilon v(x_1)]\partial_\epsilon K_\epsilon\\
&=v(x_1)K_\epsilon
+\frac{t}{h_t^2}[x_1+\epsilon v(x_1)]\\
&\quad\times\Bigl([y-(1-t)x_0-t(x_1+\epsilon v(x_1))]
\cdot v(x_1)\Bigr)K_\epsilon.
\end{align*}
Taking norms, applying Cauchy--Schwarz to the inner product, and using
$|\epsilon|\le1$, we obtain
\begin{align*}
&\left\|\partial_\epsilon\!\left([x_1+\epsilon v(x_1)]K_\epsilon\right)\right\|\\
&\quad\le \|v(x_1)\|K_\epsilon
+\frac{t}{h_t^2}
\bigl(\|x_1\|+\|v(x_1)\|\bigr)\|v(x_1)\|\\
&\qquad\quad\times
\|y-(1-t)x_0-t(x_1+\epsilon v(x_1))\|K_\epsilon\\
&\quad\le C_t\left(
\|v(x_1)\|+\|x_1\|\|v(x_1)\|+\|v(x_1)\|^2
\right).
\end{align*}
The last inequality follows because a Gaussian density is bounded, as is
its product with the distance from its mean. Applying
$ab\le(a^2+b^2)/2$ gives
\begin{equation*}
\|v(x_1)\|\le\frac{1+\|v(x_1)\|^2}{2},
\qquad
\|x_1\|\|v(x_1)\|
\le\frac{\|x_1\|^2+\|v(x_1)\|^2}{2}.
\end{equation*}
Thus, after enlarging $C_t$,
\begin{equation*}
\left\|\partial_\epsilon\!\left([x_1+\epsilon v(x_1)]K_\epsilon\right)\right\|
\le C_t\bigl(1+\|x_1\|^2+\|v(x_1)\|^2\bigr).
\end{equation*}
Similarly, $|\partial_\epsilon K_\epsilon|\le C_t\|v(x_1)\|$.
These bounds are integrable under the assumed second moments, so dominated
convergence justifies differentiation under both integrals.

Evaluating the derivatives at $\epsilon=0$, and abbreviating the kernel
arguments and $d\gamma_n=d\gamma_n(x_0,x_1)$, gives
\begin{align*}
N'(0)
&=\int v(x_1)K_0\,d\gamma_n\\
&\quad+\frac{t}{h_t^2}\int
x_1\bigl([y-(1-t)x_0-tx_1]\cdot v(x_1)\bigr)
K_0\,d\gamma_n,\\
D'(0)
&=\frac{t}{h_t^2}\int
\bigl([y-(1-t)x_0-tx_1]\cdot v(x_1)\bigr)
K_0\,d\gamma_n.
\end{align*}

The conditional distribution of the endpoints given $Y_t=y$ is
\begin{equation*}
\gamma_n(dx_0,dx_1\mid Y_t=y)
=
\frac{K_0(y\mid x_0,x_1)}{D(0)}\,d\gamma_n(x_0,x_1).
\end{equation*}
Under this conditioning, $y-(1-t)X_0-tX_1=h_tZ$. Therefore,
\begin{align*}
\frac{N'(0)}{D(0)}
&=\mathbb E[v(X_1)\mid Y_t=y]
+\frac{t}{h_t}\mathbb E[X_1(Z\cdot v(X_1))\mid Y_t=y],\\
\frac{D'(0)}{D(0)}
&=\frac{t}{h_t}\mathbb E[Z\cdot v(X_1)\mid Y_t=y].
\end{align*}
Finally, applying the quotient rule and using
$N(0)/D(0)=\bar m_t^{(n)}(y)$,
\begin{align*}
\left.\partial_\epsilon \bar m_{t,\epsilon}^{(n)}(y)\right|_0
&=\frac{N'(0)}{D(0)}
-\bar m_t^{(n)}(y)\frac{D'(0)}{D(0)}\\
&=\mathbb E[v(X_1)\mid Y_t=y]\\
&\quad+\frac{t}{h_t}\Bigl(
\mathbb E[X_1(Z\cdot v(X_1))\mid Y_t=y]\\
&\qquad\qquad
-\mathbb E[X_1\mid Y_t=y]\,
\mathbb E[Z\cdot v(X_1)\mid Y_t=y]
\Bigr)\\
&=\mathbb E[v(X_1)\mid Y_t=y]
+\frac{t}{h_t}\operatorname{Cov}\!\left(
X_1,Z\cdot v(X_1)\mid Y_t=y
\right),
\end{align*}
which proves the claim.
\end{proof}

Taking $\gamma_n=p_0\otimes q_n$ and
$v=-\nabla_{W_2}\mathcal F[q_n]$ recovers the main-text theorem whenever
the stated moment condition holds.

\subsection{Kabsch Alignment and the Induced Coupling}
\label{app:kabsch-coupling}

Let $g=(g_1,\ldots,g_N)$ and $x=(x_1,\ldots,x_N)$ be centered configurations in $\mathbb R^{3N}$. Kabsch alignment finds a rotation
\[
R^\star(g,x)\in\underset{R\in SO(3)}{\arg\min}\;
\frac12\sum_{i=1}^N\|Rg_i-x_i\|^2,
\]
where $SO(3)=\{R\in\mathbb R^{3\times3}:R^\top R=I,\ \det R=1\}$. We define the aligned configuration by $A(g,x)=(R^\star(g,x)g_i)_{i=1}^N$.

To construct the aligned coupling, we independently draw $G\sim p_0$ and $X_1\sim q_n$, then set $X_0=A(G,X_1)$. In our ALA2 experiments, $p_0$ is the empirical distribution of configurations obtained from high-temperature molecular dynamics. The resulting pair $(X_0,X_1)$ defines the coupling $\gamma_n$, and the bridge observation is
\[
Y_t=(1-t)X_0+tX_1+h_tZ,
\qquad Z\sim\mathcal N(0,I),
\]
with $Z$ independent of $G,X_1$. Here $p_0$ describes the unaligned seed $G$; the aligned sample $X_0$ need not have the same marginal distribution.

In the endpoint-transport derivative, the already aligned source $X_0$
is held fixed while $X_1$ is perturbed; alignment is not recomputed during
that infinitesimal perturbation.

To derive the Jacobian $D_xA(g,x)$, we first review basic facts about the Lie algebra of $SO(3)$.

\begin{lemma}[Derivative of a rotation]
Let $R:(-\delta,\delta)\to SO(3)$ be differentiable. Then there exists $\Omega\in T_I SO(3)$ such that
\[
R'(0)=\Omega R(0),
\qquad
T_I SO(3)=\{\Omega\in\mathbb R^{3\times3}:\Omega^\top=-\Omega\}.
\]
\end{lemma}

\begin{proof}
Differentiating $R(\epsilon)R(\epsilon)^\top=I$ at $\epsilon=0$ gives
\[
R'(0)R(0)^\top+R(0)R'(0)^\top=0.
\]
Thus $\Omega:=R'(0)R(0)^\top$ is skew-symmetric and belongs to $T_I SO(3)$. Multiplying by $R(0)$ yields $R'(0)=\Omega R(0)$.
\end{proof}

Every $\Omega\in T_I SO(3)$ can be written uniquely as
\[
\Omega=[\omega]_\times:=
\begin{pmatrix}
0&-\omega_3&\omega_2\\
\omega_3&0&-\omega_1\\
-\omega_2&\omega_1&0
\end{pmatrix},
\qquad \omega\in\mathbb R^3.
\]
Direct multiplication shows that $\Omega a=\omega\times a$. Consequently, for $a_i(\epsilon)=R(\epsilon)g_i$, the preceding lemma gives
\[
a_i'(0)=\Omega R(0)g_i=\omega\times a_i(0).
\]

\begin{lemma}[Optimality condition for alignment]
Let $R^\star$ minimize $\frac12\sum_i\|Rg_i-x_i\|^2$ over $SO(3)$, and set $a_i=R^\star g_i$. Then
\[
\sum_i a_i\times x_i=0.
\]
\end{lemma}

\begin{proof}
Fix $u\in\mathbb R^3$ and consider the rotated positions $b_i(s)=\exp(s[u]_\times)a_i$, where $\exp$ denotes the matrix exponential. These satisfy $b_i(0)=a_i$ and $b_i'(0)=u\times a_i$.

Since $a$ is optimally aligned, $E(s):=\frac12\sum_i\|b_i(s)-x_i\|^2$ has a minimum at $s=0$. Therefore,
\[
\begin{aligned}
0=E'(0)
&=\sum_i(a_i-x_i)\cdot(u\times a_i)\\
&=-\sum_i x_i\cdot(u\times a_i)\\
&=-u\cdot\sum_i a_i\times x_i.
\end{aligned}
\]
As this holds for every $u\in\mathbb R^3$, we conclude that
\[
\sum_i a_i\times x_i=0.
\]
\end{proof}

\begin{theorem}[Jacobian of Kabsch alignment]
Fix centered configurations $g,x$, and suppose the optimal rotation $R^\star(g,\cdot)$ is unique and differentiable in a neighborhood of $x$. Write $a_i=R^\star(g,x)g_i$, and assume that
\[
H:=\sum_{i=1}^N\bigl[(a_i^\top x_i)I-a_ix_i^\top\bigr]
\]
is invertible. Then, for any centered perturbation $\delta x$,
\[
\bigl[D_xA(g,x)\,\delta x\bigr]_i
=
\left(H^{-1}\sum_{j=1}^N a_j\times\delta x_j\right)\times a_i.
\]
Equivalently, defining $T\omega=(\omega\times a_i)_{i=1}^N$, we have $D_xA(g,x)=TH^{-1}T^\top$.
\end{theorem}

\begin{proof}
Let $R(\epsilon)=R^\star(g,x+\epsilon\delta x)$ and $a_i(\epsilon)=R(\epsilon)g_i$. By the rotation derivative lemma, there exists $\omega\in\mathbb R^3$ such that
\[
a_i'(0)=\omega\times a_i.
\]

For each sufficiently small $\epsilon$, the configuration $a(\epsilon)$ is optimally aligned to $x+\epsilon\delta x$. The optimality condition therefore gives
\[
\sum_i a_i(\epsilon)\times(x_i+\epsilon\delta x_i)=0.
\]
Differentiating at $\epsilon=0$ yields
\[
0=\sum_i\left[(\omega\times a_i)\times x_i
+a_i\times\delta x_i\right].
\]
Using the BAC--CAB identity \cite{jackson_classical_1999},
\[
\begin{aligned}
(\omega\times a_i)\times x_i
&=a_i(x_i^\top\omega)-(a_i^\top x_i)\omega\\
&=\bigl[a_ix_i^\top-(a_i^\top x_i)I\bigr]\omega.
\end{aligned}
\]
Substituting and rearranging, we obtain
\[
H\omega=\sum_i a_i\times\delta x_i.
\]
Since $H$ is invertible,
\[
\omega=H^{-1}\sum_i a_i\times\delta x_i.
\]
Substituting into $[D_xA(g,x)\,\delta x]_i=a_i'(0)=\omega\times a_i$ proves the stated formula.

Finally, the scalar triple-product identity gives
\[
\langle T\omega,\delta x\rangle
=\sum_i(\omega\times a_i)\cdot\delta x_i
=\omega\cdot\sum_i a_i\times\delta x_i.
\]
Thus $T^\top\delta x=\sum_i a_i\times\delta x_i$, and we conclude that
\[
D_xA(g,x)=TH^{-1}T^\top.
\]
\end{proof}

\begin{corollary}[Positivity of the bridge Jacobian]
\label{cor:kabsch-positive-jacobian}
Under the assumptions of the preceding theorem, $H\succ0$ and
$D_xA(g,x)=TH^{-1}T^\top\succeq0$. Consequently, for $0<t<1$,
\[
J_t=tI+(1-t)D_xA(g,x)\succeq tI\succ0.
\]
\end{corollary}
\begin{proof}
The optimality condition implies that $\sum_i a_ix_i^\top$ is symmetric,
so $H$ is symmetric. For any $u\in\mathbb R^3$, let
$\Omega v=u\times v$ and consider
$E(s)=\frac12\sum_i\|e^{s\Omega}a_i-x_i\|^2$.
The alignment minimum at $s=0$ gives
\[
0\leq E''(0)
=\sum_i\left[(a_i^\top x_i)\|u\|^2
-(u^\top a_i)(x_i^\top u)\right]
=u^\top H u.
\]
Thus $H\succeq0$; its assumed invertibility gives $H\succ0$.
The remaining claims follow from $D_xA=TH^{-1}T^\top$.
\end{proof}

\subsection{Stein's Identity and the Mean Estimator}
\label{app:mean-stein}

\begin{proposition}[Mean estimator under the aligned coupling]
Let $G\sim p_0$, $X_1\sim q_n$, and $Z\sim\mathcal N(0,I)$ be independent. For fixed $t\in(0,1)$ and $h_t>0$, define
\[
Y_t=(1-t)A(G,X_1)+tX_1+h_tZ,
\qquad
J_t=tI+(1-t)D_{X_1}A(G,X_1).
\]
Assume $q_n$ has a differentiable positive density, $A(G,\cdot)$ is differentiable almost surely, and the integrability and vanishing-boundary conditions required for the following applications of Stein's identity hold. Then
\[
\begin{aligned}
\mathcal M_t[q_n](y)
&:=\mathbb E\!\left[
\nabla\log\pi(X_1)-\nabla\log q_n(X_1)
\,\middle|\,Y_t=y
\right]\\
&=\mathbb E\!\left[
\nabla\log\pi(X_1)+\frac{J_t^\top Z}{h_t}
\,\middle|\,Y_t=y
\right]
\end{aligned}
\]
for almost every $y$ under the distribution of $Y_t$.
\end{proposition}

\begin{proof}
Let $\varphi:\mathbb R^d\to\mathbb R$ be a smooth, compactly supported test function. Holding $G$ and $Z$ fixed, we apply the vector Stein identity (\Cref{eqn:stein-identity}) in $X_1$ to
\[
f(x)=\varphi\!\left((1-t)A(G,x)+tx+h_tZ\right).
\]
Applying \Cref{eqn:stein-identity} with $r=q_n$ gives
\[
\mathbb E_{X_1}\!\left[
f(X_1)\nabla\log q_n(X_1)+\nabla_x f(X_1)
\right]=0.
\]
By the chain rule,
\[
\nabla_x f(X_1)
=
\underbrace{\left[tI+(1-t)D_{X_1}A(G,X_1)\right]^\top}_{J_t^\top}
\nabla\varphi(Y_t).
\]
Substituting and averaging over $G,Z$, we obtain
\[
\mathbb E\!\left[
\varphi(Y_t)\nabla\log q_n(X_1)
+J_t^\top\nabla\varphi(Y_t)
\right]=0.
\]

For standard Gaussian noise, whose score is $-Z$, the general
Stein identity in \Cref{eqn:stein-identity} specializes to
\begin{equation}\label{eqn:gaussian-stein-identity}
\mathbb E[Z\xi(Z)]=\mathbb E[\nabla\xi(Z)]
\end{equation}
for sufficiently regular scalar test functions $\xi$ satisfying
the integrability and vanishing-boundary conditions.

Next, applying the Gaussian vector Stein identity (\Cref{eqn:gaussian-stein-identity}) in $Z$, with $G,X_1$ fixed, gives
\[
\mathbb E_Z\!\left[Z\varphi(Y_t)\mid G,X_1\right]
=h_t\,\mathbb E_Z\!\left[\nabla\varphi(Y_t)\mid G,X_1\right].
\]
Since $J_t$ does not depend on $Z$, multiplying by $J_t^\top/h_t$ and averaging yields
\[
\mathbb E[J_t^\top\nabla\varphi(Y_t)]
=\frac1{h_t}\mathbb E[\varphi(Y_t)J_t^\top Z].
\]
Combining the two identities,
\[
\mathbb E\!\left[
\varphi(Y_t)
\left(\nabla\log q_n(X_1)+\frac{J_t^\top Z}{h_t}\right)
\right]=0.
\]
As this holds for every such $\varphi$, we conclude that
\[
\mathbb E[\nabla\log q_n(X_1)\mid Y_t=y]
=-\frac1{h_t}\mathbb E[J_t^\top Z\mid Y_t=y].
\]
Substitution into the definition of $\mathcal M_t[q_n]$ proves the result.
\end{proof}

For independent endpoints, $A(G,x)=G$, so $J_t=tI$, recovering the estimator $\nabla\log\pi(X_1)+tZ/h_t$. For Kabsch alignment, the preceding Jacobian theorem gives $J_t=tI+(1-t)TH^{-1}T^\top$.

\subsection{Covariance Estimators under Independent Coupling}\label{apx:cov_expansion}
Fix $t\in(0,1)$ with $h_t>0$. Suppose $X_1\sim q_n$, $X_0\sim p_0$ and
$Z\sim\mathcal{N}(0,I_d)$ are independent, and write
$s_{q_n}=\nabla\log q_n$ and $s_\pi=\nabla\log\pi$.
Let $\phi$ be the standard Gaussian density on $\mathbb{R}^d$, and set
\begin{align}
  Y &= tX_1+(1-t)X_0+h_tZ,
  & \bar m(y) &= \mathbb E[X_1\mid Y=y],\\
  F(y,x) &= f_\theta(y)\cdot\bigl(x-\bar m(y)\bigr),
  & a(x) &= \frac{h_t}{2t}s_\pi(x).
\end{align}
Here $f_\theta:\mathbb R^d\to\mathbb R^d$ is the vector field being fitted.
Using the main-text definition of $\mathcal C_t[q_n]$, we estimate its
expected inner product with $f_\theta$:
\begin{align}
  \mathcal C_t[f_\theta]
  &:=\mathbb E\!\left[f_\theta(Y)\cdot\mathcal C_t[q_n](Y)\right]
  \notag\\
  &=\frac{t}{h_t}\mathbb E\!\left[
    f_\theta(Y)\cdot
    \operatorname{Cov}\!\left(
      X_1,Z\cdot\bigl(s_\pi(X_1)-s_{q_n}(X_1)\bigr)\mid Y
    \right)
  \right].
  \label{eq:covariance-contribution}
\end{align}
All expectations below average over the sampled variables unless conditioning
is shown. We assume sufficient differentiability, absolute integrability of
the expectations involved, and vanishing boundary terms in integration by parts.
The mean $m$ and the covariance are defined under the original Gaussian bridge law.

\begin{theorem}[Unbiased covariance estimator]
\label{thm:tilted-covariance}
The function,
\begin{align}
  \widetilde\phi(z\mid x)
  &=\frac{\lVert z+a(x)\rVert^2}{d+\lVert a(x)\rVert^2}\,\phi(z),
  \label{eq:tilted-density}
\end{align}
is a probability density. For any coupling of $(Z,\widetilde Z)$ with conditional
marginals $\phi$ and $\widetilde\phi(\cdot\mid X_1)$ given $(X_0,X_1)$, set
$\widetilde Y=tX_1+(1-t)X_0+h_t\widetilde Z$. Then,
\begin{align}
  \widehat{\mathcal C}_t[f_\theta]
  &=\frac{t}{h_t}f_\theta(Y)\cdot Z
  \notag\\
  &\quad+\frac{t^2}{h_t^2}\bigl(d+\lVert a(X_1)\rVert^2\bigr)
    \bigl[F(\widetilde Y,X_1)-F(Y,X_1)\bigr]
  \label{eq:covariance-estimator}
\end{align}
satisfies $\mathbb E[\widehat{\mathcal C}_t[f_\theta]]=\mathcal C_t[f_\theta]$.
\end{theorem}

\begin{proof}
Expanding out the covariance term we get
\begin{align}
  \mathcal C_t[f_\theta]
  &=\frac{t}{h_t}\mathbb{E}\!\left[
  \underbrace{f_\theta(Y)\cdot(X_1-\bar m(Y))}_{F(Y,X_1)}
  \bigl(Z\cdot(s_\pi(X_1)-s_{q_n}(X_1))\bigr)\right]
  \notag\\
  &=\frac{t}{h_t}\mathbb{E}[FZ\cdot s_\pi(X_1)]
    -\frac{t}{h_t}\mathbb{E}[FZ\cdot s_{q_n}(X_1)].
\end{align}

For the term involving the unknown endpoint score, hold $X_0,Z$ fixed
and apply the vector Stein identity (\Cref{eqn:stein-identity}) with $r=q_n$
and scalar test function $\xi(x)=F(tx+(1-t)X_0+h_tZ,x)$.
Taking the dot product with the fixed vector $Z$ and averaging over $X_0,Z$ gives the first equality below.
 Here $\nabla_{X_1}$ includes the dependence
through $Y$, while $\partial_{X_1}F(Y,X_1)$ holds $Y$ fixed.
\begin{align}
  \mathbb{E}[ F(Y, X_1) Z \cdot s_{q_{n}}(X_1) ]
  &= -\mathbb{E}[\langle\nabla_{X_1}F(Y,X_1),Z\rangle] \\
                                                       &= - \mathbb{E}[ \langle \partial_{X_1} F(Y, X_1) + t \partial_{Y} F(Y, X_1) , Z \rangle ] \\ 
                                                       &= - \mathbb{E}[ \langle f_{\theta}(Y) + t \partial_{Y}[ f_{\theta}(Y)\cdot (X_1 - \bar m(Y)) ], Z \rangle ] \\
                                                       &= - \mathbb{E}[ \langle f_{\theta}(Y), Z \rangle  \\
                                                       &+ t \langle \nabla_{Y}[f_{\theta}(Y)\cdot (X_1 - \bar m(Y))], Z \rangle] \\
                                                       &= -  \mathbb{E}[ \langle f_{\theta}(Y), Z \rangle  \\ 
                                                       &+ \frac{t}{h_t} \langle \nabla_{Z}[F(Y, X_1)], Z \rangle ] \qquad \text{using $\nabla_Z F = h_t\nabla_Y F$}  
\end{align}
For the next step, hold $X_0,X_1$ fixed and write $F(Z)$ for
$F(Y(t,X_1,X_0,Z),X_1)$.

Apply the Gaussian vector Stein identity (\Cref{eqn:gaussian-stein-identity})
with scalar test function $\xi(z)=z_iF(z)$, and take its $i$th component to obtain
\begin{align}
  \mathbb{E}[Z_{i}^{2} F(Z)] &= \mathbb{E}[\nabla_{Z_{i}}[Z_{i} F(Z)] ] \\ 
                             &= \mathbb{E}[F(Z) + Z_{i} \nabla_{Z_{i}} F(Z)] 
\end{align}
Rearranging and summing over $i=1,\ldots,d$ yields
\begin{align}
  \mathbb{E}[\langle Z, \nabla_{Z} F \rangle] &= \mathbb{E}[ (\Vert Z \Vert^2 - d) F(Z) ] 
\end{align}
Substituting this above we get 
\begin{align}
   \mathbb{E}[ F(Y, X_1) Z \cdot s_{q_{n}}(X_1) ]  &=   -  \mathbb{E}[ \langle f_{\theta}(Y), Z \rangle \\ 
                                                         &+ \frac{t}{h_{t}} (\Vert Z \Vert^2 - d) F(Y, X_1)]  
\end{align} 
Therefore, combining with the target-score term gives
\begin{align}
\mathcal C_t[f_\theta]
&=\frac{t}{h_t}\mathbb{E}\!\left[f_\theta(Y)\cdot Z\right]
\notag\\
&\quad+\frac{t}{h_t}\mathbb{E}\!\left[
  F(Y,X_1)\underbrace{\left(
    Z\cdot s_\pi(X_1)
    +\frac{t}{h_t}\bigl(\lVert Z\rVert^2-d\bigr)
\right)}_{\text{High variance term}}
\right].
\notag
\end{align}
For $t>0$ and $h_t>0$, write $s=s_\pi(X_1)$. Completing the square gives
\begin{align} 
  Z \cdot s + \frac{t}{h_t}( \Vert Z \Vert^2 - d) = \frac{t}{h_t} \bigg \Vert Z + \frac{h_t}{2t} s \bigg \Vert^2 - \frac{h_t}{4t} \Vert s \Vert^2 - \frac{t}{h_t} d
\end{align}
Substituting this above gives
\begin{align}
\mathcal C_t[f_\theta]
&=\frac{t}{h_t}\mathbb{E}\!\left[f_\theta(Y)\cdot Z\right]
\notag\\
&\quad+\frac{t^2}{h_t^2}\mathbb{E}\!\left[
  F(Y,X_1)
  \left\lVert Z+\frac{h_t}{2t}s_\pi(X_1)\right\rVert^2
\right]
\notag\\
&\quad-\frac14\mathbb{E}\!\left[
  F(Y,X_1)\lVert s_\pi(X_1)\rVert^2
\right]
-\frac{t^2d}{h_t^2}\mathbb{E}\!\left[F(Y,X_1)\right].
\end{align}
\paragraph{Change of measure step}
Let $\phi$ be the standard Gaussian density on $\mathbb{R}^d$.
For a fixed vector $a$,
\begin{align}
  \mathbb{E}_{Z\sim\phi}\!\left[\lVert Z+a\rVert^2\right]
  &= d+\lVert a\rVert^2.
\end{align}
Now fix $X_0=x_0$ and $X_1=x_1$, and set
\begin{align}
  a(x_1) &= \frac{h_t}{2t}s_\pi(x_1),
  &
  \widetilde\phi(z\mid x_1)
  &= \frac{\lVert z+a(x_1)\rVert^2}
          {d+\lVert a(x_1)\rVert^2}\,\phi(z).
\end{align}
The preceding Gaussian identity shows that $\widetilde\phi(\cdot\mid x_1)$
is a normalized probability density. Draw
$\widetilde Z\mid X_1\sim\widetilde\phi(\cdot\mid X_1)$ and define
\begin{align}
  \widetilde Y &= tX_1+(1-t)X_0+h_t\widetilde Z.
\end{align}
The endpoint variables retain their original independent laws
$X_1\sim q_n$ and $X_0\sim p_0$; only the conditional noise law changes.
The change of measure and the law of iterated expectation give
\begin{align}
&\mathbb{E}\!\left[
  F(Y,X_1)
  \left\lVert Z+\frac{h_t}{2t}s_\pi(X_1)\right\rVert^2
\right]
\notag\\
&\quad=
\mathbb{E}_{X_0,X_1}\!\left[
  \bigl(d+\lVert a(X_1)\rVert^2\bigr)
  \mathbb{E}_{\widetilde Z\sim\widetilde\phi(\cdot\mid X_1)}
  \!\left[F(\widetilde Y,X_1)\right]
\right]
\notag\\
&\quad=
\mathbb{E}\!\left[
  \left(d+\frac{h_t^2}{4t^2}\lVert s_\pi(X_1)\rVert^2\right)
  F(\widetilde Y,X_1)
\right].
\end{align}
The two subtracted terms from completing the square combine as
\begin{align}
&-\frac14\mathbb{E}\!\left[
  F(Y,X_1)\lVert s_\pi(X_1)\rVert^2
\right]
-\frac{t^2d}{h_t^2}\mathbb{E}\!\left[F(Y,X_1)\right]
\notag\\
&\quad=
-\frac{t^2}{h_t^2}\mathbb{E}\!\left[
  \bigl(d+\lVert a(X_1)\rVert^2\bigr)F(Y,X_1)
\right].
\end{align}
Substituting both identities into the covariance expression therefore yields
\begin{align}
\mathcal C_t[f_\theta]
&=\frac{t}{h_t}\mathbb{E}\!\left[f_\theta(Y)\cdot Z\right]
\notag\\
&\quad+\frac{t^2}{h_t^2}\mathbb{E}\!\left[
  \bigl(d+\lVert a(X_1)\rVert^2\bigr)F(\widetilde Y,X_1)
\right]
-\frac{t^2}{h_t^2}\mathbb{E}\!\left[
  \bigl(d+\lVert a(X_1)\rVert^2\bigr)F(Y,X_1)
\right]
\notag\\
&=\frac{t}{h_t}\mathbb{E}\!\left[f_\theta(Y)\cdot Z\right]
\notag\\
&\quad+\frac{t^2}{h_t^2}\mathbb{E}\!\left[
  \bigl(d+\lVert a(X_1)\rVert^2\bigr)
  \bigl(F(\widetilde Y,X_1)-F(Y,X_1)\bigr)
\right].
\end{align}

This is exactly $\mathbb E[\widehat{\mathcal C}_t[f_\theta]]$ for the estimator
in \Cref{eq:covariance-estimator}. Only the conditional noise marginals enter
the two expectations, so the identity holds for any coupling with those marginals.
\end{proof}

Their coupling affects variance but not the expectation. It remains to draw
from the tilted density. The following lemma gives an exact sampler.

\begin{lemma}[Sampling the tilted density]
\label{lem:tilted-sampling}
Fix $a\in\mathbb R^d$. Draw independent $G\sim\mathcal N(0,I_d)$ and
$H_1,H_2\sim\mathcal N(0,1)$. Independently choose
\begin{align}
  V&=\begin{cases}
    G,&\text{with probability }\displaystyle\frac{\lVert a\rVert^2}{d+\lVert a\rVert^2},\\[1ex]
    \displaystyle\sqrt{1+\frac{H_1^2+H_2^2}{\lVert G\rVert^2}}\,G,
      &\text{otherwise},
  \end{cases}
  \label{eq:symmetric-sampler}
\end{align}
then set
\begin{align}
  \widetilde Z&=\begin{cases}
    V,&\text{with probability }\displaystyle\frac12+
       \frac{V\cdot a}{\lVert V\rVert^2+\lVert a\rVert^2},\\[1ex]
    -V,&\text{otherwise}.
  \end{cases}
  \label{eq:tilted-sign}
\end{align}
Then $\widetilde Z$ has density
$\widetilde\phi(z\mid a)=\lVert z+a\rVert^2\phi(z)/(d+\lVert a\rVert^2)$.
\end{lemma}

\begin{proof}
Start by choosing a pair $\{z,-z\}$ using the symmetrized density below.
Then conditional on $\{z,-z\}$ choose one of these, this corresponds to:
\begin{align}
  P(\text{choose } z \mid \pm z) &= \frac{\widetilde \phi(z)}{\widetilde \phi(z) + \widetilde \phi(-z)} \\ 
                                 &= \frac{\lVert z + a \rVert^2 }{\lVert z + a \rVert^2 + \lVert z - a \rVert^2} \\ 
                                 &= \frac12 + \frac{z \cdot a}{\lVert z \rVert^2 + \lVert a \rVert^2}
\end{align}

\begin{align}
   \frac12 \left( \widetilde\phi(z\mid a) + \widetilde\phi(-z \mid a) \right) 
  &= \frac{\lVert z \rVert^2 + \lVert a \rVert^2}
          {d+\lVert a\rVert^2}\,\phi(z) 
\end{align}

Now consider the symmetrized density $s(z \mid a) = \frac12 \left( \widetilde \phi(z \mid a) + \widetilde \phi(- z \mid a) \right)$. Note that
\begin{align}
  s(z \mid a) P(\text{choose }z \mid \pm z)
  &= \frac{\widetilde\phi(z\mid a)}
          {\widetilde\phi(z\mid a)+\widetilde\phi(-z\mid a)}
     \cdot\frac12\left(\widetilde\phi(z\mid a)+\widetilde\phi(-z\mid a)\right)
  \notag\\
  &= \frac12\widetilde\phi(z\mid a).
\end{align}
But we can also output $z$ by first drawing $-z$ and choosing its opposite.
Adding both contributions, the output density at $z$ is
\begin{align}
&s(z\mid a)P(\text{choose }z\mid\pm z)
+s(-z\mid a)P(\text{choose }z\mid\pm z)
\notag\\
&\qquad=\frac12\widetilde\phi(z\mid a)
       +\frac12\widetilde\phi(z\mid a)
       =\widetilde\phi(z\mid a).
\end{align}
Thus, it suffices to sample from the symmetrized density and then choose the
sign with the probability above. For fixed $a$, write $g(z)=s(z\mid a)$, so
\begin{align}
  g(z) &= \frac{\lVert z \rVert^2 + \lVert a \rVert^2}{d + \lVert a \rVert^2} \phi(z) \\ 
       &= \frac{\lVert a \rVert^2}{\lVert a \rVert^2 + d} \phi(z) + \frac{d}{\lVert a \rVert^2 + d} \frac{\lVert z \rVert^2}{d} \phi(z)
\end{align}

This is just a mixture of two densities $\phi(z)$ and $d^{-1}\lVert z\rVert^2\phi(z)$.
The second is normalized since $\mathbb{E}_{\phi}[\lVert Z\rVert^2]=d$.
Multiplying the Gaussian density by $\lVert z\rVert^2$ changes its radial
density from one proportional to $r^{d-1}e^{-r^2/2}$ to one proportional to
$r^{d+1}e^{-r^2/2}$. This is just the Gaussian radius in $d+2$ dimensions.

Recall that sampling from a Gaussian may be done by sampling the radius with
density proportional to $r^{d-1}e^{-r^2/2}$ and independently drawing a direction
from $\mathcal{U}(S^{d-1})$, namely
\begin{align}
  G &= RU, \qquad U\sim\mathcal{U}(S^{d-1}), \qquad R\perp U,\\
  p_R(r) &\propto r^{d-1}e^{-r^2/2}, \qquad r>0.
\end{align}
Equivalently, given $G\sim\mathcal{N}(0,I_d)$, take $R=\lVert G\rVert$ and
$U=G/\lVert G\rVert$. The Gaussian density depends only on the radius, and the
volume element in polar coordinates contributes the factor $r^{d-1}$.
This is why the direction is uniform and independent of the radius.

Multiplying the Gaussian density by $\lVert z\rVert^2$ changes only its radial
density. So to sample from a density proportional to $\lVert z\rVert^2\phi(z)$,
draw two independent one dimensional standard normals $H_1,H_2$, also
independent of $G$, and do
\begin{align}
  \widetilde R
  &= \sqrt{G_1^2+\cdots+G_d^2+H_1^2+H_2^2},
  \qquad H_1,H_2\sim\mathcal{N}(0,1),\\
  \widetilde G
  &= \widetilde R\,U
   = \sqrt{1+\frac{H_1^2+H_2^2}{\lVert G\rVert^2}}\,G.
\end{align}
Now $\widetilde R^2$ is the sum of $d+2$ independent squared standard normals,
so $\widetilde R$ has the required radial density. Also, $U$ is independent
of $(R,H_1,H_2)$ and hence of $\widetilde R$. Thus $\widetilde G$ has density
$d^{-1}\lVert z\rVert^2\phi(z)$. It is still a vector in $\mathbb{R}^d$;
the extra two scalars change only its length.

Thus \Cref{eq:symmetric-sampler} draws $V$ from $s(\cdot\mid a)$.
The sign selection in \Cref{eq:tilted-sign} then gives a draw from
$\widetilde\phi(\cdot\mid a)$, as shown above.
\end{proof}

For the estimator, take $a=a(X_1)$. Using $Z=G$ for the original Gaussian noise
and this $\widetilde Z$ for the tilted noise gives a coupling with the required
marginals. We can therefore substitute $Y=tX_1+(1-t)X_0+h_tG$ and
$\widetilde Y=tX_1+(1-t)X_0+h_t\widetilde Z$ into
\Cref{eq:covariance-estimator}. The construction is exact; it does not by
itself guarantee lower variance.

\subsection{Covariance Estimator under a General Coupling}
\label{app:covariance-general-coupling}

\paragraph{Stein's identity under a general coupling.}
Let $G\sim p_0$, $X_1\sim q_n$, and $Z\sim\mathcal N(0,I_d)$ be independent.
For fixed $t\in(0,1)$ and $h_t>0$, define
\[
Y_t=(1-t)A(G,X_1)+tX_1+h_tZ,
\qquad
J_t=tI+(1-t)D_{X_1}A(G,X_1).
\]
Write $s_{q_n}=\nabla\log q_n$ and
$F(y,x)=f_\theta(y)\cdot(x-\bar m(y))$, where
$\bar m(y)=\mathbb E[X_1\mid Y_t=y]$.
We assume sufficient smoothness and integrability for the following
integrations by parts, with vanishing boundary terms.

Holding $G,Z$ fixed, we apply Stein's identity in $X_1$ to the vector field
$F(Y_t,X_1)Z$. The chain rule gives
\[
\nabla_{X_1}F(Y_t,X_1)
=f_\theta(Y_t)+J_t^\top\nabla_yF(Y_t,X_1),
\]
where the first term comes from differentiating the explicit $X_1$ in $F$,
and the second accounts for its dependence through $Y_t$. Therefore,
\begin{align*}
\mathbb E[F(Y_t,X_1)Z\cdot s_{q_n}(X_1)]
&=-\mathbb E\!\left[
Z\cdot\nabla_{X_1}F(Y_t,X_1)
\right]\\
&=-\mathbb E[f_\theta(Y_t)\cdot Z]
-\mathbb E[\nabla_yF(Y_t,X_1)\cdot J_tZ]\\
&=-\mathbb E[f_\theta(Y_t)\cdot Z]
-\frac1{h_t}\mathbb E[\nabla_ZF(Y_t,X_1)\cdot J_tZ],
\end{align*}
using $\nabla_ZF(Y_t,X_1)=h_t\nabla_yF(Y_t,X_1)$.

Next, hold $G,X_1$ fixed and apply Gaussian Stein's identity to
$F(Y_t,X_1)J_tZ$. Since $J_t$ does not depend on $Z$, the product rule yields
\begin{align*}
\mathbb E[F(Y_t,X_1)Z^\top J_tZ]
&=\mathbb E\!\left[
\nabla_Z\cdot\bigl(F(Y_t,X_1)J_tZ\bigr)
\right]\\
&=\mathbb E\!\left[
\nabla_ZF(Y_t,X_1)\cdot J_tZ
+F(Y_t,X_1)\operatorname{tr}J_t
\right].
\end{align*}
Rearranging and substituting into the preceding expression, we obtain
\begin{align}
\mathbb E[F(Y_t,X_1)Z\cdot s_{q_n}(X_1)]
&=-\mathbb E[f_\theta(Y_t)\cdot Z]
\notag\\
&\quad-\frac1{h_t}\mathbb E\!\left[
F(Y_t,X_1)
\bigl(Z^\top J_tZ-\operatorname{tr}J_t\bigr)
\right].
\label{eq:covariance-general-stein}
\end{align}
For independent coupling, $A(G,x)=G$ and $J_t=tI$, recovering the factor
$t(\lVert Z\rVert^2-d)$ from the preceding proof.

\paragraph{Completing the square and changing measure.}
We retain the covariance convention from the main text:
\begin{equation}
\mathcal C_t[f_\theta]
:=\frac{t}{h_t}\mathbb E\!\left[
F(Y_t,X_1)Z\cdot\bigl(s_\pi(X_1)-s_{q_n}(X_1)\bigr)
\right].
\label{eq:general-covariance-contribution}
\end{equation}
The following construction requires the symmetric part
$S_t=(J_t+J_t^\top)/2$ to be positive definite almost surely, in addition
to the integration-by-parts and integrability assumptions above. This
holds for independent coupling and for nondegenerate Kabsch alignment
(Corollary~\ref{cor:kabsch-positive-jacobian}). The mean $m$ remains the
conditional mean under the original Gaussian bridge.

\begin{theorem}[Covariance estimator under a general coupling]
\label{thm:general-covariance-estimator}
Define
\[
a=\frac{h_t}{2}S_t^{-1}s_\pi(X_1),
\qquad
\kappa=\operatorname{tr}S_t+a^\top S_ta.
\]
Let $\phi$ denote the standard Gaussian density. Conditional on $G,X_1$,
the function,
\begin{equation}
\widetilde\phi(z\mid G,X_1)
=\frac{(z+a)^\top S_t(z+a)}{\kappa}\,\phi(z),
\label{eq:general-quadratic-tilt}
\end{equation}
is a probability density. For any coupling of $Z,\widetilde Z$ with
conditional marginals $\phi,\widetilde\phi$ given $G,X_1$, set
$\widetilde Y_t=(1-t)A(G,X_1)+tX_1+h_t\widetilde Z$. Then,
\begin{equation}
\widehat{\mathcal C}_t[f_\theta]
=\frac{t}{h_t}f_\theta(Y_t)\cdot Z
+\frac{t\kappa}{h_t^2}
\bigl[F(\widetilde Y_t,X_1)-F(Y_t,X_1)\bigr]
\label{eq:general-covariance-estimator}
\end{equation}
satisfies $\mathbb E[\widehat{\mathcal C}_t[f_\theta]]=\mathcal C_t[f_\theta]$.
\end{theorem}

\begin{proof}
Substituting \Cref{eq:covariance-general-stein} into
\Cref{eq:general-covariance-contribution}, and using
$Z^\top J_tZ=Z^\top S_tZ$ and
$\operatorname{tr}J_t=\operatorname{tr}S_t$, gives
\begin{align}
\mathcal C_t[f_\theta]
&=\frac{t}{h_t}\mathbb E[f_\theta(Y_t)\cdot Z]\notag\\
&\quad+\frac{t}{h_t^2}\mathbb E\!\left[
F(Y_t,X_1)
\left(Z^\top S_tZ+h_tZ\cdot s_\pi(X_1)-\operatorname{tr}S_t\right)
\right].
\end{align}
Since $2S_ta=h_ts_\pi(X_1)$, completing the square yields,
\[
Z^\top S_tZ+h_tZ\cdot s_\pi(X_1)-\operatorname{tr}S_t
=(Z+a)^\top S_t(Z+a)-\kappa.
\]
Conditional on $G,X_1$, the matrix $S_t$ and vector $a$ are fixed, and
\[
\mathbb E_Z[(Z+a)^\top S_t(Z+a)]
=\operatorname{tr}S_t+a^\top S_ta=\kappa.
\]
Together with $S_t\succ0$, this establishes that
\Cref{eq:general-quadratic-tilt} is a probability density. Changing
measure conditional on $G,X_1$, then averaging, gives
\[
\mathbb E\!\left[F(Y_t,X_1)(Z+a)^\top S_t(Z+a)\right]
=\mathbb E[\kappa F(\widetilde Y_t,X_1)].
\]
Combining these identities,
\begin{align*}
\mathcal C_t[f_\theta]
&=\frac{t}{h_t}\mathbb E[f_\theta(Y_t)\cdot Z]\notag\\
&\quad+\frac{t}{h_t^2}\mathbb E\!\left[
\kappa\bigl(F(\widetilde Y_t,X_1)-F(Y_t,X_1)\bigr)\right],
\end{align*}
which is the expectation of \Cref{eq:general-covariance-estimator}.
\end{proof}

\begin{lemma}[Sampling a quadratic Gaussian tilt]
\label{lem:general-quadratic-tilt-sampling}
Fix $S\succ0$ and $a\in\mathbb R^d$, and write
$S=Q\operatorname{diag}(\lambda_1,\ldots,\lambda_d)Q^\top$,
$b=Q^\top a$, and $\kappa=\operatorname{tr}S+a^\top Sa$.
Choose an index $I=i$ with probability
$w_i=\lambda_i(1+b_i^2)/\kappa$. Conditional on $I=i$, draw $U_i$
from the one-dimensional density,
\[
\frac{(u+b_i)^2}{1+b_i^2}\,\phi_1(u),
\]
using Lemma~\ref{lem:tilted-sampling} in dimension one. Draw the remaining
coordinates independently from $\mathcal N(0,1)$ and independently of
$U_i$. Then $\widetilde Z=QU$ has density
$\kappa^{-1}(z+a)^\top S(z+a)\phi(z)$.
\end{lemma}
\begin{proof}
The weights sum to one because
$\sum_i\lambda_i(1+b_i^2)=\kappa$. The mixture density of $U$ is
\[
\sum_i w_i\frac{(u_i+b_i)^2}{1+b_i^2}\prod_j\phi_1(u_j)
=\frac{\sum_i\lambda_i(u_i+b_i)^2}{\kappa}\,\phi(u).
\]
The orthogonal transformation $z=Qu$ preserves the standard Gaussian
density and turns the numerator into $(z+a)^\top S(z+a)$.
\end{proof}

For independent coupling, $S_t=tI$, so
$a=h_ts_\pi(X_1)/(2t)$ and $\kappa=t(d+\|a\|^2)$, recovering
\Cref{eq:covariance-estimator}. For Kabsch alignment,
$S_t=J_t=tI+(1-t)TH^{-1}T^\top\succ0$. The Stein identity above also
holds for other differentiable couplings, but the positive-definiteness
condition is required for this particular completed-square tilt.

\subsection{Training Objective Derivation}
\label{app:regression-objective}
Fix an outer iteration $n$ and hold the denoiser $\bar m^{(n)}$ fixed.
Write $r_t^{(n)}=\mathcal M_t[q_n]+\lambda_t\mathcal C_t[q_n]$ as in
\Cref{eqn:wasabi-correction-direction}. Assume the relevant second moments
and estimator expectations are finite, and that the score-free identities
hold, including the conditional-mean assumption on $\bar m^{(n)}$.
Expanding the ideal regression loss gives
\begin{align}
\mathcal J_n(\theta)
&:=\frac12\mathbb E\!\left[\|f_\theta(Y_t,t)-r_t^{(n)}(Y_t)\|^2\right]\notag\\
&=\frac12\mathbb E\|f_\theta(Y_t,t)\|^2
-\mathbb E\!\left[f_\theta(Y_t,t)\cdot\mathcal M_t[q_n](Y_t)\right]\notag\\
&\quad-\mathbb E\!\left[\lambda_t f_\theta(Y_t,t)\cdot\mathcal C_t[q_n](Y_t)\right]
+K_n,
\label{eq:regression-loss-expansion}
\end{align}
where $K_n=\tfrac12\mathbb E\|r_t^{(n)}(Y_t)\|^2$ is independent of $\theta$.
The expectation averages over bridge time, source and endpoint samples,
and the original Gaussian noise, as specified in Section~\ref{sec:wasabi-practical-algorithm}.

Set $A_t=-\nabla U(\widetilde X_1)+(t/h_t)Z$.
By \Cref{eqn:mean-estimator-score-free} and the tower property,
\[
\mathbb E[f_\theta(Y_t,t)\cdot\mathcal M_t[q_n](Y_t)]
=\mathbb E[f_\theta(Y_t,t)\cdot A_t].
\]
For each fixed $t$, applying \Cref{thm:covariance-tilted-main} with
$\varphi=f_\theta(\cdot,t)$ gives
\[
\mathbb E[\lambda_t f_\theta(Y_t,t)\cdot\mathcal C_t[q_n](Y_t)]
=\mathbb E[\lambda_t\widehat{\mathcal C}_t[f_\theta(\cdot,t)]],
\]
where the right-hand expectation also includes the tilted noise.
Substituting these two identities into \Cref{eq:regression-loss-expansion}
yields $\mathcal J_n(\theta)=\mathcal L_n(\theta)+K_n$, with
$\mathcal L_n$ given in \Cref{eqn:wasabi-practical-loss}.
Thus the two population objectives have the same minimizers;
this equivalence does not remove finite-sample or denoiser-approximation errors.

\clearpage
\subsection{Extended WASABI Algorithm}
\label{app:wasabi-extended}

\begin{algorithm}[H]
\caption{WASABI with replay, optional alignment, and annealing}
\label{alg:wasabi-extended}
\algrenewcommand{\algorithmicrequire}{\textbf{Inputs:}}
\begin{algorithmic}[1]
\Require Energy $U$; sampleable prior $p_0$; $K$ initial endpoints;
refresh count $R\leq K$; noise scale $B$; iterations $N$;
schedules $\eta_t,\lambda_t,\beta_n$; coupling map $A$.
\State Initialize denoiser parameters $\alpha$ and correction parameters $\theta$.
\State Initialize the ring buffer
$\mathcal D_0=\{(x^{(k)},\nabla U(x^{(k)}))\}_{k=1}^{K}$.
\State Set $A(g,x)=g$ without alignment and $\beta_n=1$ without annealing.
\For{$n=0,\ldots,N-1$}
    \State \textbf{Set target score:} $s_{\pi_n}(x)=-\beta_n\nabla U(x)$.
    \State \textbf{Fit denoiser:} sample minibatches of
    $(X_1,\nabla U(X_1))\sim\mathcal D_n$, independent $G\sim p_0$,
    bridge times $t$, and $Z\sim\mathcal N(0,I)$. Construct
    \Statex \hspace{\algorithmicindent}\hspace{\algorithmicindent}
    $X_0=A(G,X_1)$, $\quad h_t=B\sqrt{t(1-t)}$,
    \Statex \hspace{\algorithmicindent}\hspace{\algorithmicindent}
    $Y_t=(1-t)X_0+tX_1+h_tZ$, and fit
    \Statex \hspace{\algorithmicindent}\hspace{\algorithmicindent}
    $\displaystyle
    \alpha_n\approx\arg\min_\alpha
    \mathbb E_{\mathcal D_n}\!\left[\|m_\alpha(Y_t,t)-X_1\|^2\right].$
    \State \textbf{Fit correction:} freeze $\alpha_n$; construct training
    pairs as above and compute
    \Statex \hspace{\algorithmicindent}\hspace{\algorithmicindent}
    $J_t=tI+(1-t)D_{X_1}A(G,X_1)$.
    \Statex \hspace{\algorithmicindent} Using paired Gaussian and tilted noise, fit
    \Statex \hspace{\algorithmicindent}\hspace{\algorithmicindent}
    $\displaystyle
    \begin{aligned}
    \theta_n\approx\arg\min_\theta\;\mathbb E_{\mathcal D_n}\!\Bigg[
    &\frac12\|f_\theta(Y_t,t)\|^2
    -\lambda_t\widehat{\mathcal C}_{t,n}[f_\theta(\cdot,t)]\\
    &-f_\theta(Y_t,t)\cdot
    \left(s_{\pi_n}(X_1)+\frac{J_t^\top Z}{h_t}\right)
    \Bigg].
    \end{aligned}$
    \State \textbf{Update sampler:}
    $\widetilde m_t^{(n+1)}(y)\gets
    m_{\alpha_n}(y,t)+\eta_t f_{\theta_n}(y,t)$.
    \State \textbf{Generate fresh endpoints:} simulate the updated sampler
    to obtain $R$ configurations $x$ and evaluate their gradients $\nabla U(x)$.
    \State \textbf{Refresh buffer:} replace the oldest $R$ entries of
    $\mathcal D_n$ with the fresh pairs $(x,\nabla U(x))$ to obtain $\mathcal D_{n+1}$.
\EndFor
\State \Return The sampler driven by $\widetilde m^{(N)}$.
\end{algorithmic}
\end{algorithm}

Here $\widehat{\mathcal C}_{t,n}$ denotes the general-coupling covariance
estimator from Appendix~\ref{app:covariance-general-coupling}, evaluated
using the current target score $s_{\pi_n}$ and the frozen denoiser
$m_{\alpha_n}$. The subscript $\mathcal D_n$ indicates that endpoints
are drawn uniformly from the replay buffer; the expectations also average
over prior configurations, bridge times, and noise.

The prior $p_0$ need only support sampling: prior configurations may be
generated on demand or drawn from a stored sample bank, without evaluating
the prior density or score. In the ALA2 experiments, we use a bank of
configurations generated by a molecular dynamics simulation of the ALA2
Hamiltonian (Appendix~\ref{apx:ala2-hamiltonian}) at $1{,}000\,\mathrm K$.
This prior bank is distinct from the replay buffer, which stores generated
endpoints together with their energy gradients and is partially refreshed
at each outer iteration. When annealing, the cached gradients are rescaled
by the current factor $\beta_n$.

Retaining endpoints from earlier samplers makes replay-buffer training an
approximation to the population iteration. This reuse follows prior
work~\citep{liu2025asbs,blessing2026bms}, and empirically we find that it
produces samples that closely match the target distributions defined by
the potentials in Section~\ref{sec:experiments}.

\clearpage
\section{Experiment Details}
\label{app:experiment-details}
\subsection{DW-4 Evaluation}
\label{app:dw4-evaluation}

We compare ASBS, damped ASBS, BMS, damped BMS, and WASABI using a
$16\times4$ EGNN under matched training settings. Training stops at
1,000 outer rounds for every method. For each seed, we pool 128 fresh
endpoints from each of 16 checkpoints, saved every 25 rounds from 625
through 1,000, giving 2,048 endpoints. Each run is evaluated against one
random 2,048-frame subsample of the public test split. Configuration
$W_2$ uses the released assignment convention~\citep{blessing2026bms}.
Energy $W_2$ uses every endpoint, without an energy cutoff.

Table~\ref{tab:dw4-preliminary} reports means and standard deviations
across six seeds, except for damped BMS, where seed 1 is excluded from
both metrics after divergence in this window (energy $W_2=2{,}449$).
The retained five seeds give configuration $W_2=1.48\pm0.21$ and energy
$W_2=2.72\pm0.58$. Including all six seeds gives $1.46\pm0.20$ and
$410\pm999$, respectively.

The pair-distance plot in Figure~\ref{fig:dw4-pair-distances} pools all six
pairwise distances from 10,000 WASABI and 10,000 reference configurations.
It is a separate illustrative sample set, not the pooled checkpoint
samples used for the table. The dashed lines mark the minima of the
pairwise potential.

\subsection{DW-4 and LJ-13 potentials}
\label{apx:particle-potentials}
For both systems, $r_{ij}=\|x_i-x_j\|$ denotes the distance between
particles $i$ and $j$.

\paragraph{Double-well system (DW-4).}
The four particles lie in two dimensions and interact through
\begin{equation}
  U_{\mathrm{DW-4}}(x)
  =\sum_{i<j}\left[0.9(r_{ij}-4)^4-4(r_{ij}-4)^2\right].
  \label{eq:dw4-potential}
\end{equation}

\paragraph{Lennard--Jones cluster (LJ-13).}
The 13 particles lie in three dimensions. Their pairwise interactions
are supplemented by quadratic confinement around the center of mass:
\begin{equation}
  U_{\mathrm{LJ-13}}(x)
  =2\sum_{i<j}\left(r_{ij}^{-12}-2r_{ij}^{-6}\right)
  +\frac12\sum_{i=1}^{13}\|x_i-x_{\mathrm{COM}}\|^2,
  \qquad x_{\mathrm{COM}}=\frac1{13}\sum_{i=1}^{13}x_i.
  \label{eq:lj13-potential}
\end{equation}

\subsection{Alanine-dipeptide potential and training setup}
\label{apx:ala2-hamiltonian}
Following the alanine-dipeptide benchmark in \citep{blessing2026bms},
we use ff96 parameters
with OBC1 implicit solvent and an additional chirality restraint.
Let \(x=(r_1,\ldots,r_n)\in(\mathbb{R}^3)^n\) denote the atomic
positions. The configurational potential is
\begin{equation}
\begin{aligned}
U(x) &= U_{\mathrm{ff96}}(x)
      + U_{\mathrm{OBC1}}(x)
      + U_{\mathrm{chirality}}(x),\\
U_{\mathrm{ff96}}(x)
    &= U_{\mathrm{bond}}(x)
     + U_{\mathrm{angle}}(x)
     + U_{\mathrm{torsion}}(x)
     + U_{\mathrm{LJ}}(x)
     + U_{\mathrm{Coulomb}}(x),\\
U_{\mathrm{OBC1}}(x)
    &= U_{\mathrm{GB}}(x)+U_{\mathrm{SA}}(x).
\end{aligned}
\end{equation}
The nonzero bonded coefficients of this Hamiltonian define the index sets
\begin{equation}
\begin{aligned}
E &= \{e=(i,j): k_e^{\mathrm b}\ne 0\},\\
A &= \{a=(i,j,k): k_a^{\mathrm a}\ne 0\},\\
T &= \{\tau=(i,j,k,l): V_{\tau n}\ne 0\text{ for some }n\}.
\end{aligned}
\label{eq:ala2-interaction-sets}
\end{equation}
Here \(j\) is the central atom of an angle triple, and \(n\) indexes the
Fourier components of a dihedral. For this alanine-dipeptide topology,
\(A\) coincides with the triples formed by pairs of bonds in \(E\)
sharing a central atom.

\begin{table}[H]
\setlength{\belowcaptionskip}{11pt}
\caption{Energy terms in the alanine-dipeptide Hamiltonian. The formulas display
functional forms; ff96 exclusions and special 1--4 parameters are included in
the starred nonbonded sums. Coefficients include any thermal rescaling used to
define \(U\).}
\label{tab:ala2-hamiltonian}
\centering
\normalsize
\setlength{\tabcolsep}{3pt}
\renewcommand{\arraystretch}{1.35}
\begin{tabular}{@{}>{\raggedright\arraybackslash}p{0.17\linewidth}>{\raggedright\arraybackslash}p{0.51\linewidth}>{\raggedright\arraybackslash}p{0.22\linewidth}@{}}
\toprule
\textbf{Term} & \textbf{Energy expression} & \textbf{Physics} \\
\midrule
\(U_{\mathrm{bond}}\) &
\(\displaystyle \frac12\sum_{e\in E}k_e^{\mathrm b}(\ell_e-\ell_e^0)^2\) &
Covalent bond stretching. \\
\(U_{\mathrm{angle}}\) &
\(\displaystyle \frac12\sum_{a\in A}k_a^{\mathrm a}(\alpha_a-\alpha_a^0)^2\) &
Valence-angle bending. \\
\(U_{\mathrm{torsion}}\) &
\(\displaystyle \sum_{\tau\in T}\sum_n V_{\tau n}[1+\cos(n\phi_\tau-\gamma_{\tau n})]\) &
Rotation barriers about bonds. \\
\(U_{\mathrm{LJ}}\) &
\(\displaystyle 4\sum_{i<j}^{\star}\epsilon_{ij}(\rho_{ij}^{12}-\rho_{ij}^{6})\) &
Steric repulsion and dispersion. \\
\(U_{\mathrm{Coulomb}}\) &
\(\displaystyle k_e\sum_{i<j}^{\star}\frac{q_iq_j}{r_{ij}}\) &
Direct charge interactions. \\
\(U_{\mathrm{GB}}\) &
\(\displaystyle -\frac{k_e\Delta_\epsilon}{2}\sum_{i,j}\frac{q_iq_j}{f_{ij}^{\mathrm{GB}}}\) &
Implicit-solvent electrostatics. \\
\(U_{\mathrm{SA}}\) &
\(\displaystyle 4\pi\gamma_{\mathrm{SA}}\sum_i(a_i+a_s)^2(a_i/R_i)^6\) &
Nonpolar solvation cost. \\
\(U_{\mathrm{chirality}}\) &
\(\displaystyle \frac{k_\chi}{2}\sum_s[\max(d_s-\tau,0)]^2\) &
Molecular handedness restraint. \\
\bottomrule
\end{tabular}

\end{table}
Here \(r_{ij}=\|r_i-r_j\|\), \(\rho_{ij}=\sigma_{ij}/r_{ij}\), and
\(\Delta_\epsilon=\epsilon_{\mathrm{in}}^{-1}-\epsilon_{\mathrm{out}}^{-1}\).
The generalized-Born pair distance is
\begin{equation*}
f_{ij}^{\mathrm{GB}}=
\left[r_{ij}^2+R_iR_j\exp\!\left(-\frac{r_{ij}^2}{4R_iR_j}\right)\right]^{1/2}.
\end{equation*}
The \(R_i\) are OBC1 effective Born radii, \(a_i\) are atomic radii, and
\(a_s\) is the solvent probe radius. For each restrained dihedral \(s\),
\(d_s=|\operatorname{wrap}(\chi_s-\chi_s^0)|\) is its wrapped angular distance
from the preferred chirality, with tolerance \(\tau\).

\paragraph{Bond--angle split readout.}
The sets \(E\) and \(A\) defined by the nonzero bond and angle terms
index the bond lengths and angle cosines used in our readout. We define
the local bonded-coordinate map
\begin{equation}
\begin{aligned}
c(x) &=
\bigl(
\{\ell_{ij}(x)\}_{(i,j)\in E},
\{\cos\alpha_{ijk}(x)\}_{(i,j,k)\in A}
\bigr),\\
\ell_{ij}(x) &= \|r_i-r_j\|,\\
\cos\alpha_{ijk}(x) &=
\frac{(r_i-r_j)\cdot(r_k-r_j)}
{\|r_i-r_j\|\,\|r_k-r_j\|}.
\end{aligned}
\label{eq:ala2-bond-angle-map}
\end{equation}
All Jacobians below are taken in the orthonormal zero-centroid coordinate
basis used by the model.
Thus \(c(x)\) uses the known connectivity and current positions, but
neither evaluates the Hamiltonian nor uses its force constants or preferred
bond lengths. With \(J(x)=Dc(x)\), a small displacement satisfies
\(c(x+\varepsilon v)=c(x)+\varepsilon J(x)v+O(\varepsilon^2)\).
Directions in \(\ker J(x)\) leave the selected bond lengths and angles
unchanged to first order.

To distinguish these motions in the corrector, we use the regularized map
\begin{equation}
P_\delta(x)=J(x)^\top
\bigl(J(x)J(x)^\top+\delta I\bigr)^{-1}J(x),
\qquad \delta=10^{-4},
\label{eq:ala2-bash-map}
\end{equation}
\begin{samepage}
and combine the two PaiNN vector heads as
\begin{equation}
f_\theta(x,t)=
\sigma_{\mathrm{stiff}}P_\delta(x)H_{\mathrm{stiff}}(x,t)
+\sigma_{\mathrm{soft}}\bigl(I-P_\delta(x)\bigr)
H_{\mathrm{soft}}(x,t).
\label{eq:ala2-bash-readout}
\end{equation}
\end{samepage}
\begin{samepage}
We call this the Bond--Angle Split Heads (BASH) readout. The reported ALA2
runs use a separate PaiNN backbone for each head, trained through their
sum. The split is
evaluated at the current noisy configuration. Because \(P_\delta\) is
regularized, this is not an exact orthogonal decomposition and does not
fix bond lengths or angles. In particular, \(c(x)\) does not encode
handedness; the separate chirality restraint acts through the target
energy and score.
\end{samepage}

\paragraph{Training setup and prior preparation.}
\label{apx:ala2-practical}
The reported sweeps share a 4,096-endpoint replay buffer and 256 fresh samples
per outer iteration (6.25\%). We linearly interpolate inverse temperature
from 1,000\,K to 300\,K over the first 1,500 of 4,000 outer iterations:
\begin{equation}
\beta_n=\frac{1-\alpha_n}{k_{\mathrm B}(1{,}000\,\mathrm K)}
       +\frac{\alpha_n}{k_{\mathrm B}(300\,\mathrm K)},
\qquad
\alpha_n=\min\!\left\{\frac{n}{1{,}500},1\right\},
\label{eq:ala2-inverse-temperature-schedule}
\end{equation}
where $n\geq0$ is the outer-iteration index and
$\beta_n=(k_{\mathrm B}T_n)^{-1}$. The target remains at 300\,K thereafter.
Replay samples are reweighted as the temperature changes.
The fixed prior bank is separate from this evolving replay buffer.
It contains 20,000 configurations from the 1,000\,K state of a seven-temperature
CUDA REMD run; the full run, including all temperatures and equilibration,
took approximately 62.5 minutes. An independent chain supplies the
20,000-frame reference in Figure~\ref{fig:ala2-results}.
A separate local benchmark generated 25,000 configurations at 1,000\,K
in approximately five minutes on an Apple M5 Max using OpenMM's CPU
platform with one thread. This timing includes system setup, minimization,
100,000 equilibration steps and 2,500,000 production steps at 1\,fs,
recording every 100 steps; plotting is excluded. The 20k-frame local-prior
budget in Table~\ref{tab:ala2-comparison} uses 2,000,000 production steps
and the same equilibration length.

We compared this locally generated 20k-frame bank with the 1,000\,K states
of two independent CUDA REMD chains, using 20,000 frames per bank and the
same $50\times50$ angle grid. Local MD versus the two REMD banks gave
$D_{\rm JS}=0.0309$ and $0.0307$, compared with $0.0296$ between the two
REMD banks. The discrepancy between the local MD and REMD backbone-angle
distributions is comparable to the observed chain-to-chain sampling variability.
Based on this agreement, we do not expect replacing the REMD-derived prior
with the locally generated high-temperature MD prior to materially affect
the reported results. The reported training runs used the REMD-derived bank.

\paragraph{Ramachandran maps.}
The maps in Figure~\ref{fig:ala2-results} show relative free energy
$\Delta F/(k_{\mathrm B}T)=-\log(p/p_{\max})$, where $p$ is the histogram-bin
probability and $p_{\max}$ is its maximum within each panel. Free-energy
zeros are therefore set separately, and gray bins contain no observed
samples. The positive-$\phi$ region contains 10 of 20,000 WASABI samples
(0.05\%) and 78 of 20,000 reference frames (0.39\%).

Figure~\ref{fig:ala2-results} compares 20,000 WASABI samples with 20,000
frames from an independent REMD simulation. We show seed 1 for
$s_\eta=0.10$, $\lambda_{\rm cov}=0.25$, which had the lowest JS divergence
against this reference among the three seeds.

\clearpage
\subsection{Alanine-dipeptide ablations and computational cost}
\label{apx:ala2-results}
\label{sec:ablations-practical}

\begin{figure}[H]
\centering
\includegraphics[width=\linewidth]{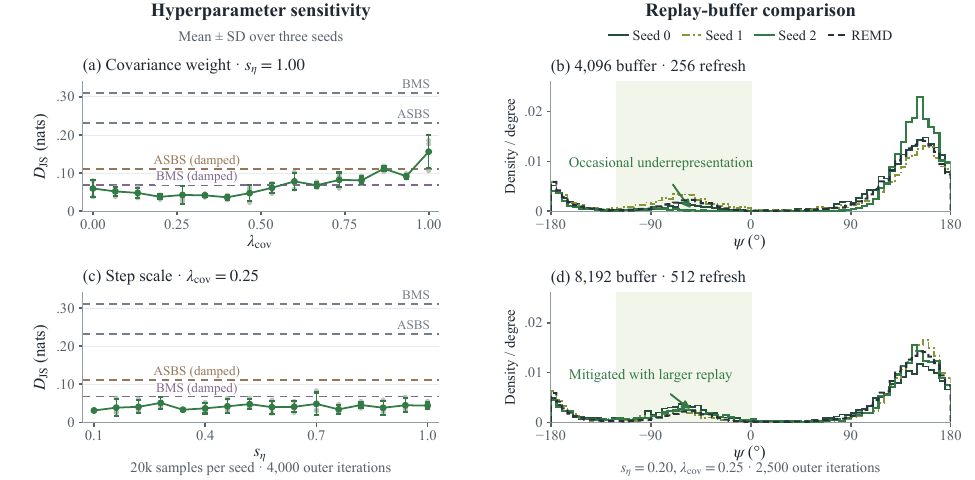}
\caption{\textbf{ALA2 ablations and replay comparison.}
\textbf{(a,c)} JS mean $\pm$ SD over three seeds; gray dots show individual
scores and dashed lines published BMS/ASBS means~\citep{blessing2026bms}.
\textbf{(b,d)} Matched seeds with smaller and larger replay settings;
dashed curves show REMD. Arrows highlight mitigation of seed 2's
central-peak underrepresentation. Both capacity and refresh increase.
The sweeps and replay comparison use different checkpoints and sample
counts; evaluation protocols are described below.}
\label{fig:ala2-ablation}
\label{fig:ala2-ablation-buffer}
\end{figure}

\paragraph{Influence of hyperparameters.}
We separately vary the covariance weight and update step scale, holding
$s_\eta=1$ fixed in the covariance sweep and $\lambda_{\mathrm{cov}}=0.25$
fixed in the step-scale sweep. Each sweep includes 16 settings evaluated
over three training seeds, for 96 runs in total
(Figure~\ref{fig:ala2-ablation}(a,c), Table~\ref{tab:ala2-full}).
Covariance weights below $0.5$ generally give lower JS divergence, with
the lowest mean at $\lambda_{\mathrm{cov}}=0.4$. Performance varies less
across the tested step scales, with the lowest mean at $s_\eta=0.10$.

\begin{table}[tbp]
\setlength{\belowcaptionskip}{11pt}
\caption{Covariance-weight and step-scale sweeps. Values are mean $\pm$ SD
over three seeds; bold marks the lowest mean in each sweep.}
\label{tab:ala2-full}
\centering\normalsize
\renewcommand{\arraystretch}{1.12}
\setlength{\tabcolsep}{6pt}
\begin{tabular*}{\linewidth}{@{\extracolsep{\fill}}rc@{\hspace{1.4em}}rc@{}}
\toprule
\multicolumn{2}{c}{\textbf{Covariance sweep}} & \multicolumn{2}{c}{\textbf{Step-scale sweep}}\\
\multicolumn{2}{c}{$s_\eta=1.00$ fixed} & \multicolumn{2}{c}{$\lambda_{\rm cov}=0.25$ fixed}\\
\cmidrule(lr){1-2}\cmidrule(l){3-4}
$\lambda_{\rm cov}$ & $D_{\rm JS}\downarrow$ & $s_\eta$ & $D_{\rm JS}\downarrow$\\
\midrule
$0.0000$ & $0.0595 \pm 0.0212$ & $0.10$ & \textbf{0.0312\,$\pm$\,0.0023} \\
$0.0667$ & $0.0520 \pm 0.0125$ & $0.16$ & $0.0387 \pm 0.0218$ \\
$0.1333$ & $0.0479 \pm 0.0137$ & $0.22$ & $0.0408 \pm 0.0173$ \\
$0.2000$ & $0.0377 \pm 0.0061$ & $0.28$ & $0.0512 \pm 0.0167$ \\
$0.2667$ & $0.0426 \pm 0.0237$ & $0.34$ & $0.0332 \pm 0.0020$ \\
$0.3333$ & $0.0416 \pm 0.0047$ & $0.40$ & $0.0370 \pm 0.0135$ \\
$0.4000$ & \textbf{0.0365\,$\pm$\,0.0061} & $0.46$ & $0.0422 \pm 0.0180$ \\
$0.4667$ & $0.0475 \pm 0.0216$ & $0.52$ & $0.0479 \pm 0.0139$ \\
$0.5333$ & $0.0613 \pm 0.0135$ & $0.58$ & $0.0400 \pm 0.0173$ \\
$0.6000$ & $0.0782 \pm 0.0224$ & $0.64$ & $0.0409 \pm 0.0145$ \\
$0.6667$ & $0.0681 \pm 0.0079$ & $0.70$ & $0.0485 \pm 0.0299$ \\
$0.7333$ & $0.0822 \pm 0.0194$ & $0.76$ & $0.0342 \pm 0.0111$ \\
$0.8000$ & $0.0808 \pm 0.0102$ & $0.82$ & $0.0456 \pm 0.0089$ \\
$0.8667$ & $0.1117 \pm 0.0073$ & $0.88$ & $0.0384 \pm 0.0179$ \\
$0.9333$ & $0.0918 \pm 0.0062$ & $0.94$ & $0.0449 \pm 0.0185$ \\
$1.0000$ & $0.1562 \pm 0.0435$ & $1.00$ & $0.0441 \pm 0.0095$ \\
\bottomrule
\end{tabular*}

\end{table}

\paragraph{Replay-buffer size and mode coverage.}
A low JS divergence can coexist with underrepresentation of a smaller peak
in the backbone-angle distribution. Figure~\ref{fig:ala2-ablation}(b,d)
illustrates this behavior in the central peak of the $\psi$ distribution.
Increasing the replay capacity from 4,096 to 8,192 and the number of fresh
endpoints per iteration from 256 to 512 improves coverage in the
highlighted seed, although peak populations remain variable across the
five runs evaluated with the larger replay setting. Both replay capacity
and refresh count are changed in this comparison.

\paragraph{Evaluation.}
Following \citet{blessing2026bms}, we compute JS divergence in nats using a
$50\times50$ grid of backbone angles. The hyperparameter sweeps use 20,000
generated samples per seed at outer iteration 4,000, evaluated against
the one-million-frame validation set released with
\href{https://arxiv.org/abs/2208.01893}{FAB}
(\href{https://zenodo.org/records/6993124}{reference dataset}).
Values are reported as mean $\pm$ standard deviation over three seeds.
Evaluating the same selected checkpoints with 1,024,000 generated samples
per seed reduces JS from $0.0312\pm0.0023$ to $0.0254\pm0.0033$
(Table~\ref{tab:ala2-comparison}), consistent with reduced sampling noise
in the empirical histograms. The replay comparison uses matched seeds
0--2 at iteration 2,500, with $s_\eta=0.20$, $\lambda_{\mathrm{cov}}=0.25$,
4,096 generated samples per seed, and an independent 20,000-frame REMD
reference. The plotted $\psi$ marginals use unsmoothed $5^\circ$ bins.

\begin{samepage}
\paragraph{Computational cost.}
We count matrix multiplications in PaiNN's message, update, and readout
blocks, including backpropagation, with each multiply-add counted as two
FLOPs. Applying a linear map from $d_{\mathrm{in}}$ to $d_{\mathrm{out}}$
channels to $K$ feature vectors therefore costs approximately
\[
2K\,d_{\mathrm{in}}d_{\mathrm{out}}\ \text{FLOPs}.
\]
We sum these costs using the tensor dimensions in the implementation.
Other operations, including activations, geometric calculations, optimizer
updates, and molecular-energy evaluations, are excluded.
\end{samepage}

For ALA2, one evaluation of our four-layer, 64-channel PaiNN costs
76.6 MFLOPs per molecule. Sampling evaluates two correction networks and
one denoiser, totaling 229.7 MFLOPs per integration step. Each outer
iteration performs 100 correction updates and 200 denoiser updates with
batch size 64, then generates 256 fresh endpoints using 256 integration
steps. Including the antithetic and tilted queries used during training
gives 31.7 TFLOPs per outer iteration.\footnote{Counts retain repeated
frozen-denoiser evaluations before compiler optimization. Including the
10,000-update denoiser warmup and the skipped first denoiser refit gives
31.79 TFLOPs per outer iteration averaged over all 4,000 rounds.}

For BMS, we assume eight GPUs based on the paper's 16,384-sample replay
buffer and the \href{https://github.com/DenisBless/BridgeMatchingSampler}{released configuration}'s 2,048 samples per GPU
\citep{blessing2026bms}. This implies a global training batch of 512 and
2,048 fresh endpoints per outer iteration. Under these settings, we
estimate 112.9 TFLOPs per outer iteration without damping and
190.2 TFLOPs with damping. The paper does not explicitly report the GPU
count. We do not estimate ASBS training FLOPs because its corrector and
adjoint calculations require separate accounting.

\paragraph{Molecular energy and force evaluations.}
\label{apx:ala2-energy-calls}
Table~\ref{tab:ala2-comparison} reports estimated molecular-potential calls
in millions per run, including force evaluations. One endpoint scored for
both energy and force counts once. Neural sampler evaluations and optimizer
updates are not molecular-potential calls. Diagnostic and evaluation
samples, minimization, and additional energy-only reporting queries are
excluded from these nominal force-call budgets.

WASABI scores the 4,096 initial replay endpoints and 256 fresh endpoints
at each of 4,000 outer iterations, totaling
$4,096+4,000\times256=1,028,096$ calls. Inner updates reuse cached scores;
annealing rescales cached physical forces and reweights cached energies
without new potential evaluations. Charging each run for a local 20k-frame
hot bank adds 2,000,000 production and 100,000 equilibration MD steps,
yielding $3,128,096\approx3.13\times10^6$ calls. This is the budget for the
local-prior procedure, rather than a measured end-to-end cost of the
reported REMD-prior training runs.

For ASBS and BMS, we use the paper's 5,000 ALA2 outer iterations and the
inferred global refresh of 2,048 endpoints, yielding
$5,000\times2,048=10.24\times10^6$ calls. The released BMS configuration
initializes replay with the same number of endpoints as a refresh, counted
in its first outer iteration. ASBS's released upstream implementation
scores fresh adjoint-buffer endpoints and reuses cached terminal costs;
corrector fitting uses the analytic reference-diffusion score and does not
query the molecular potential. Damping adds neural evaluations, but no
additional molecular-potential calls under the same refresh schedule.
These counts estimate the published setting, rather than measure its
unreleased execution logs; the released BMS default of 7,000 iterations
would instead imply $14.336\times10^6$ calls.

The REMD row uses seven temperatures, 20,000 production cycles and 1,000
equilibration cycles, with 500 MD steps per cycle at 1\,fs. Across replicas,
this gives $7\times(20,000+1,000)\times500=73.5\times10^6$ force evaluations.
There are additionally 147,000 energy queries for exchange decisions and
initial minimization work. This row records the cost of the reference
protocol; it is not an accuracy-matched comparison with trained samplers,
and no self-comparison JS score is reported.

\end{document}